\documentclass[final,1p,times]{elsarticle}

\usepackage{graphicx}
\usepackage{mathbbold}
\usepackage{subfigure}
\usepackage{amsmath}
\usepackage{amssymb}
\usepackage{chemsym}
\usepackage{algorithm}
\usepackage{algorithmic}

\newtheorem{theorem}{Theorem}[section]

\newproof{pf}{Proof}

\journal{}

\begin{document}

\begin{frontmatter}



\title{Novel variational model for inpainting in the wavelet domain}


\author{Dai-Qiang Chen^{1}, Li-Zhi Cheng}

\address{Department of Mathematics and System, School of Sciences,
National University of Defense Technology, Changsha 410073, Hunan,
People's Republic of China} \fntext[Dai-Qiang Chen]{Corresponding
author, telephone number: $(+86)13467530489$}
\ead{chener050@sina.com}

\begin{abstract}
Wavelet domain inpainting refers to the process of recovering the
missing coefficients during the image compression or transmission
stage. Recently, an efficient algorithm framework which is called
Bregmanized operator splitting (BOS) was proposed for solving the
classical variational model of wavelet inpainting. However, it is
still time-consuming to some extent due to the inner iteration. In
this paper, a novel variational model is established to formulate
this reconstruction problem from the view of image decomposition.
Then an efficient iterative algorithm based on the split-Bregman
method is adopted to calculate an optimal solution, and it is also
proved to be convergent. Compared with the BOS algorithm the
proposed algorithm avoids the inner iteration and hence is more
simple. Numerical experiments demonstrate that the proposed method
is very efficient and outperforms the current state-of-the-art
methods, especially in the computational time.

\end{abstract}


\end{frontmatter}



\section{Introduction}\label{sec1}
In the image science, the term ``inpainting" means recovering the
missing data in images. It is an important task in various image
restoration problems including scratch removal, impulse noise
removal, zooming and so on. These mentioned applications are all
related to the inpainting in the image domain. However, in the real
storage and transmission of the digital images, the discrete
wavelet transform is a very popular method for the sparse coding.
For example, in the JPEG2000 image compression standard, the images
are formatted in the wavelet domain. Then part of the wavelet
coefficients has to be discarded or may be corrupted during the
compression and transmission process, which may lead to serious
degradation of the visual quality of the images especially when the loss
appears in the coarsest low-low subband. Therefore, in order to
improve the visual quality, the missing or corrupted coefficients
should be recovered from the known information. This generates
another important inpainting process in the wavelet domain, which is
the core problem discussed here.

Inpainting in the image domain, which uses the values of the
available pixels to fill the missing pixels, has been widely
investigated in the last two decades. Bertalmio et al. \cite{SIGGRAPH:Inpainting} introduced
the partial differential equations (PDEs) to smoothly propagate the
information from the neighboring pixels into the areas contain the
unknown pixels. Later, Chan and Shen proposed the total variation
(TV) inpainting model \cite{SIAM:TVInpainting}, which derives from the well-known
Rudin-Osher-Fatemi (ROF) model and aims at minimizing the TV norm in
the process of filling the missing pixels. Further, higher order
methods such as the curvature-driven diffusion (CDD) \cite{VCIR:CDD} and Euler's
elastica based variational model \cite{SIAM:EulerInpainting} were also applied to the
inpainting problem in order to overcome the block effects caused by
the TV model. These edge-preserved methods are unable to recover the
texture regions efficiently. Therefore, the exemplar-based texture
synthesis techniques \cite{TIP:Exemplar} \cite{SIAM:Exemplar} were developed for recovering the textured
images. The similar idea was also applied to image denoising and then
non-local mean filter was proposed by Buades et al. \cite{SIAM:Nonlocal}. Based on
this method, non-local total variation \cite{SIAM:NLTV}\cite{IET:NLTV}\cite{ECCV:NLTV} was further investigated
and applied in various image restoration problem including the
inpainting task. These non-local techniques are more suitable for
recovering the texture and fine details of images than the previous
local methods.

The wavelet domain inpainting is a completely different problem
since missing or damaged wavelet coefficients could cause
degradation widely spread in the pixel domain. Specifically, the loss
of different wavelet subbands leads to different degradation types on the
visual quality of images. For example, the loss of the high
frequencies in the coarsest subbands creates Gibbs artifacts or other
blur effects, and the loss in the coarsest low-low subband creates
regular or irregular big black squares in the image domain \cite{IPI:NLTVWavelet}. Due
to the damaged regions caused by the wavelet loss cannot be well
defined in the pixel domains, the block-based inpainting methods \cite{TIP:Interp} \cite{TIP:Filling}
such as image domain inpainting or interpolation algorithms, which
were widely used for the restoration of the JPEG compressed images
based on block DCT transforms, aren't well suitable for the
inpainting in the wavelet domain.

These novel features and challenges prompt the need of developing
new models and methods of image inpainting in the wavelet domain. An
important guiding principle is to control the regularization in the
image domain with the wavelet domain constraints, so that the
reconstruction images could retain the important geometrical features
such as the sharp edges. Inspired by this idea, the hybrid
variational models were introduced as an alternative class of
approaches. Durand et al. \cite{SIAM:WaveletTVInpainting} used the TV minimization technique to
recover the missing/corrupted wavelet coefficients. Chan et al. \cite{JMIV:WTVInpainting}
proposed both the constrained/unconstrained variational models
combining the TV regularization term and the wavelet coefficients
fitting. The authors also investigated the properties of the models
such as the existence and uniqueness. Besides, other edge-preserved
regularization term with different properties were also applied to
this problem \cite{SSVM:L0TVInpainting} \cite{AAS:plaplaceInpainting}. However, the classical total variation and other
local-gradient based methods are not suitable for images with fine
structures, details and textures. Then similarly to the inpainting in
the image domain, Zhang et al. \cite{IPI:NLTVWavelet} extended the TV wavelet inpainting
model to the corresponding non-local (NL) form, and proposed an
efficient iterative algorithm called Bregmanized operator splitting
(BOS) to solve the TV/NL-TV wavelet inpainting models.

In fact, these models discussed above are all established as special cases
of variational image restoration problem. However, the characteristic of
wavelet domain inpainting itself inspires us to consider the problem
from another view of image decomposition---
the original image is divided into two components: one is just the
known image constructed by the coefficients in the known domain, and
the other is the rest image decided by the missing coefficients.
Inspired by this viewpoint, we establish a novel variational model to
recover the missing coefficients based on TV/NL-TV prior for the
original image, and propose a fast iterative algorithm based on the
split-Bregman method.

The rest of the paper is organized as follows: several classical
wavelet inpainting models and corresponding fast algorithms are
introduced in section 2. In section 3 we propose a new variational
model for the inpainting problem in the view of image decomposition,
and investigate the corresponding fast algorithm.
In section 4 the existence of solutions of the corresponding
minimization problem are investigated and the convergence of the
proposed iterative algorithm is proved. In section 5 we focus on the
implementation of the proposed algorithm, and compare our model with
previous works. The conclusion about the proposed algorithm is given
in section 5.

\section{Related works}\label{sec2}
\setcounter{equation}{0}

Let $f\in L^{2}(\Omega)$ be the original image defined on the
bounded region $\Omega$, the standard orthogonal wavelet expansion
of $f$ is denoted by
\[
f(\alpha,x)=\sum_{j,k}\alpha_{j,k}\psi_{j,k}(x), ~~j\in \mathbb{Z},
k\in \mathbb{Z}^{2}
\]
where $x\in \Omega$, and $\alpha_{j,k}$ denotes wavelet coefficients
of $f$ at level $j$ and location $k$ under the wavelet basis
$\{\psi_{j,k}\}$. Assume that the wavelet coefficients in the index
region $I$ are known, that is,
\[
\beta_{j,k}=\begin{cases}\alpha_{j,k},\quad \ \ & (j,k)\in I,\\
  0, \quad \ \ & else.
\end{cases}
\]
The aim of wavelet domain inpainting is to reconstruct the wavelet
coefficients of $f$ using the given coefficients $\beta_{j,k}$. In
\cite{JMIV:WTVInpainting} Chan et al. investigated the TV model for the inpainting
problems. Precisely, the following constrained and unconstrained TV
inpainting model (corresponding to the noise-free and noise case)
are considered:
\begin{equation}\label{equ2.1}
\min_{\alpha_{j,k}}\textrm{TV}(f(\alpha,x)) ~~s.t.~~
\alpha_{j,k}=\beta_{j,k},
\end{equation}
and
\begin{equation}\label{equ2.2}
\min_{\alpha_{j,k}}\textrm{TV}(f(\alpha,x)) + \sum_{j,k}
\lambda_{j,k}(\alpha_{j,k}-\beta_{j,k})^{2}
\end{equation}
where $\lambda_{j,k}$ are regularization parameters, and the total
variation term is defined by
\[
\textrm{TV}(f)=\textrm{sup}\left\{\int_{\Omega}f~ \textrm{div}
\vec{\phi}~dx: \vec{\phi}\in (C_{0}^{\infty}(\Omega))^{2},
\|\vec{\phi}\|_{\infty}\leq 1\right\}.
\]
Assume that $f\in W^{1,1}(\Omega)$, then we further
have $\textrm{TV}(f)=\int_{\Omega}|\nabla f|dx$.

Let $W: L^{2}(\Omega)\rightarrow L^{2}(\Omega)$ denote the
orthogonal wavelet transform, and $P_{I}$ denote the projection
operator onto the known index set $I$, i.e.
\[
P_{I}(\alpha)_{j,k}=\begin{cases}\alpha_{j,k},\quad \ \ & (j,k)\in I,\\
  0, \quad \ \ & else.
\end{cases}
\]
Using a general convex regularization function $J(\cdot)$ instead
of the TV term, then the above minimization problems can be
rewritten as
\begin{equation}\label{equ2.3}
\min_{f}J(f) ~~s.t.~~  P_{I}Wf = \beta,
\end{equation}
and
\begin{equation}\label{equ2.4}
\min_{f}J(f) + \lambda\|P_{I}Wf -\beta\|^{2}_{2}.
\end{equation}
Let $A=P_{I}W$, then both formulations (\ref{equ2.3}) and
(\ref{equ2.4}) are special cases of the inverse restoration problem.
Therefore, a number of methods for solving the general inverse problem
such as the subgradient projection \cite{JSTSP:GPSR}, the Newton-like method \cite{ICAOS:CGM}, the
split-Bregman/Augmented Lagrangian method \cite{SIAM:SplitBregman} \cite{TIP:STV} \cite{IPI:ALNonQuadratic},
and the primal-dual method \cite{TIP:WTVPD} \cite{JMIV:PDConvergence} \cite{UCLA:FastAlgorithmTV} can also be
applied to solve the unconstrained problem (\ref{equ2.4}). On the other hand, the
constrained form is a different problem. In \cite{IPI:NLTVWavelet}, Zhang et al. used an
efficient algorithm called BOS to calculate a optimal solution of
the constrained problem (\ref{equ2.3}). The BOS was introduced in
\cite{SIAM:BOS}, which use the Bregman iteration \cite{SIAM:Bregman} to solve the equality
constrained problem (\ref{equ2.3}), i.e.
\begin{equation}\label{equ2.5}
\left\{\begin{array}{lll}f^{k+1}= \textrm{arg}\min_{f}\{\mu
J(f)+\frac{1}{2}\|Af-\beta^{k}\|^{2}\},
& ~ & ~\\
\beta^{k+1}=\beta^{k} + (\beta - A f^{k+1}), & ~ &~ ~
\end{array} \right.
\end{equation}
and the sub-minimization problem in (\ref{equ2.5}) is solved by the
proximal Forward-Backward Splitting algorithm (PFBS) \cite{SIAM:PFBS}: let
$f^{k+1,0}=f^{k}$, for $i>0$
\begin{equation}\label{equ2.6}
f^{k+1,i} = arg\min_{f}J(f) + \frac{1}{2\delta\mu}\left\|f
-(f^{k+1,i}-\delta A^{T}(Af^{k+1,i}-\beta^{k}))\right\|^{2}.
\end{equation}
In the above formulas, $\mu$ and $\delta$ are positive parameters. In
order to recover images which contain rich textures or fine structures,
Zhang et al. also extended the TV inpainting model to the non-local
TV form, where the NL-TV regularization term can be expressed as follows:
\begin{equation}\label{equ2.7}
\int_\Omega|\nabla_w f|(x)dx=\int_\Omega\sqrt{\int_\Omega
(f(y)-f(x))^2 w(x,y)dy} dx
\end{equation}
where $x,y\in \Omega$ and $w(x,y)$ denotes the weight function
related to pixels $x$ and $y$. For more details refer to \cite{SIAM:NLTV} \cite{IPI:NLTVWavelet}.

\section{The proposed inpainting model}\label{sec3}
\setcounter{equation}{0}

In this section, we consider the inpainting problem from another
view of image decomposition. For convenience the following
discussion is based on the discrete setting, that is, let
$\Omega=[1,\sqrt{n}]\times [1,\sqrt{n}]$, $f\in \mathbb{R}^{n}$ be
the original image and $W\in \mathbb{R}^{n\times n}$ be the
orthogonal wavelet transform matrix. Assume $\tilde{W}=W^{-1}$ is
the inverse wavelet transform, $\alpha_{I}$ and $\alpha_{I^{c}}$ are
vectors formed by the wavelet coefficients of the original image $f$
corresponding to the known index set $I$ and its complementary set.
Then we have
\[
f=\tilde{W} \alpha = \tilde{W}\alpha_{I}+\tilde{W}\alpha_{I^{c}} =
f_{0}+\tilde{W}\alpha_{I^{c}}.
\]
To observe the above relationship, the original image is decomposed
into two components---the observed image $f_{0}$ which is decided by
the known coefficients, and the unknown part which represents the
missing coefficients. This inspires us to estimate the missing
information directly based on certain priors about the original
image.

To this end, utilizing the regularization prior such as the TV/NL-TV
norm for the image $f$ and the above relationship, we could
establish the following minimization problem:
\begin{equation}\label{equ3.1}
\min_{f, \alpha_{m}\in C} J(f) ~~ s.t. ~~ f =
f_{0}+\tilde{W}\alpha_{m}
\end{equation}
where $J$ is the regularization function such as the TV/NL-TV norm, and the set
\[
C = \{\gamma: P_{I}(\gamma) = \textbf{0}\}.
\]
In fact, we observe that the component $f_{0}$ in the proposed model
can be substituted by any observed image $f^{*}$ such that $P_{I}
Wf^{*}=\alpha_{I}$.

Let $\iota_{C}$ be the indicator function of set $C$, i.e.
\[ \iota_{C}(\gamma)=
\begin{cases}0,\quad \ \ & \gamma\in C,\\
  + \infty, \quad \ \ & \gamma\not \in C.
\end{cases}
\]
Then the problem (\ref{equ3.1}) can be rewritten as
\begin{equation}\label{equ3.2}
\min_{f, \alpha_{m}} J(f)+\iota_{C}(\alpha_{m}) ~~ s.t. ~~ f =
f_{0}+\tilde{W}\alpha_{m}.
\end{equation}
The split-Bregman method can be used to solve the proposed model.
Considering a general constrained problem of the form
\begin{equation}\label{equ3.3}
\min_{u}\hat{J}(u) ~~ s.t. ~~ u\in \mathbb{R}^{m}, ~~M u=d
\end{equation}
where $\hat{J}$ is a closed, proper, convex function, $d\in
\mathbb{R}^{n}$, and $M \in \mathbb{R}^{n\times m}$ is a bounded
linear operator. The Bregman iteration \cite{SIAM:Bregman} for (\ref{equ3.3}) can be
reformulated as
\begin{equation}\label{equ3.4}
\left\{\begin{array}{lll}u^{k+1}=\textrm{arg}\min_{u}\left\{\hat{J}(u)+\frac{\lambda}{2}\|b^{k}+Mu-d\|^{2}\right\}
& ~ &~ ~\\ b^{k+1}=b^{k} + (Mu^{k+1} - d)& ~ &~
\end{array} \right.
\end{equation}
By identifying $(f,\alpha_{m}), J(f)+\iota_{C}(\alpha_{m}),
\tilde{W}\alpha_{m}-f$, and $-f_{0}$ as $u, \hat{J}(u), Mu$ and $d$,
the Bregman iteration for problem (\ref{equ3.2}) is
\begin{equation}\label{equ3.5}
\left\{\begin{array}{lll}(f^{k+1},
\alpha_{m}^{k+1})=\textrm{arg}\min_{f,\alpha_{m}}\left\{J(f)+\iota_{C}(\alpha_{m})
+\frac{\lambda}{2}\|b^{k}+f_{0}+\tilde{W}\alpha_{m}-f\|^{2}\right\}, & ~ &~
~\\ b^{k+1}=b^{k} + (f_{0}+\tilde{W}\alpha_{m}^{k+1}-f^{k+1}).& ~ &~
\end{array} \right.
\end{equation}
In the above iteration, the sub-minimization problem can be solved
by the alternative minimization approach, that is,
\begin{equation}\label{equ3.6}
\left\{\begin{array}{lll}
\alpha_{m}^{k+1}=\textrm{arg}\min_{\alpha_{m}}\left\{\iota_{C}(\alpha_{m})
+\frac{\lambda}{2}\|b^{k}+f_{0}+\tilde{W}\alpha_{m}-f^{k}\|^{2}\right\},& ~
&~ ~ \\ f^{k+1}=\textrm{arg}\min_{f}\left\{J(f)
+\frac{\lambda}{2}\|b^{k}+f_{0}+\tilde{W}\alpha_{m}^{k+1}-f\|^{2}\right\}.
&~ &~ ~\\
\end{array} \right.
\end{equation}

The first sub-minimization problem in (\ref{equ3.6}) has the closed
solution as follows
\[
\alpha_{m}^{k+1}=P_{C}\left(W(f^{k}-f_{0}-b^{k})\right)
\]
where $P_{C}$ denotes the projection to the set $C$, which is just
to set the coefficients on the index set $I$ to be zeros and keep
others unchanged. Moreover, for the general case of non-orthogonal
wavelet transform such as redundant transform and tight frames,
$\tilde{W}$ is not orthogonal, but still has full rank. Let
$\hat{W}$ be the matrix which is constructed by columns of
$\tilde{W}$ corresponding to the index in set $I^{c}$, then the
non-zero elements of $\alpha_{m}^{k+1}$ can be obtained by
$\hat{W}^{\dag}(f^{k}-f_{0}-b^{k})$, where $\hat{W}^{\dagger}$
denotes the general inverse of $\hat{W}$.


The second sub-minimization problem of the formula (\ref{equ3.6})
has the same form as the formula (\ref{equ2.6}), which is just the
iterative step of the inner iteration of the BOS algorithm. In this
paper we choose the regularization function $J$ to be TV/NL-TV norm.
Thus this sub-problem corresponds to the TV or NL-TV denoising model
, and it can be solved by the primal-dual or split-Bregman methods
efficiently. For more details refer to \cite{SIAM:SplitBregman}\cite{UCLA:FastAlgorithmTV}. Introducing the proximity
operator \cite{SIAM:PFBS} defined by
\[
\textrm{prox}_{J}(g)={\textrm{arg}\min}\left
\{\frac{1}{2}\|g-f\|^{2}_{2}+J(f)\right \},
\]
the solution of the second sub-minimization problem can be expressed
as $f^{k+1}=\textrm{prox}_{\frac{1}{\lambda}
J}(b^{k}+f_{0}+\tilde{W}\alpha_{m}^{k+1})$.

During the implementation of Bregman iteration (\ref{equ3.5}), it
could guarantee the convergence enough to update $f^{k+1}$ and
$\alpha_{m}^{k+1}$ once by the approach (\ref{equ3.6}). This leads
to the split-Bregman method for problem (\ref{equ3.1}), which is
summarized as Algorithm 1 below. Note that the true solution of the
proximity operator $\textrm{prox}_{\frac{1}{\lambda}J}$ cannot be obtained in
practice. However, the convergence property of the un-exact split
Bregman iteration still comes into existence while the proximity
operator is solved exactly enough, which will be verified in the
next section.

\begin{algorithm}[htb]
\caption{ wavelet inpainting algorithm based on the split-Bregman
method}
\begin{algorithmic}[1]
\REQUIRE  observed image $f_{0}$; index set $I$ on which the wavelet
coefficient information is available; wavelet transform matrix $W$;
penalty parameter $\lambda$; step size $\delta$.\\
\textbf{Output}: estimated image $f$. \\
\textbf{Initialization}: $k=0; b^{0}=0; f^{0}=0; $\\
\textbf{Iteration}: \\
~~~\textbf{Step1}:
$\alpha_{m}^{k+1}=P_{C}\left(W(f^{k}-f_{0}-b^{k})\right)$; \\
~~~\textbf{Step2}: $f^{k+1}=\textrm{prox}_{\frac{1}{\lambda}
J}(b^{k}+f_{0}+\tilde{W}\alpha_{m}^{k+1})$; \\
~~~\textbf{Step3}: $b^{k+1}=b^{k} +
(f_{0}+\tilde{W}\alpha_{m}^{k+1}-f^{k+1})$ \\
until some stopping criterion is satisfied. \\
Use the output of $f_{0}+\tilde{W}\alpha_{m}^{k}$ of the above loop
as the final estimation of $f$.
\end{algorithmic}
\end{algorithm}

Compared with the BOS algorithm in (\ref{equ2.5})-(\ref{equ2.6}),
our algorithm avoids the inner iteration for solving the
sub-minimization problem in (\ref{equ2.5}), which requires the
implementation of the discrete wavelet and inverse wavelet transforms for several times,
as shown in the formula (\ref{equ2.6}). Therefore, our method could reduce
the computation time efficiently. Besides, the step 3 in Algorithm 1 can
be rewritten as
\[
b^{k+1}=(b^{k}+f_{0}+\tilde{W}\alpha_{m}^{k+1})-f^{k+1}.
\]
Due to $b^{k}+f_{0}+\tilde{W}\alpha_{m}^{k+1}$ and $f^{k+1}$ are
both obtained in the step 2, only one implementation of wavelet and inverse wavelet
transforms is needed in each iteration.

\section{The analysis of the proposed model}\label{sec4}
\setcounter{equation}{0}

In the following, we choose the regularization function $J$ to be
TV/NL-TV norm, and then address the existence of the
solution of problem (\ref{equ3.1}). To this end, we require that the
inverse wavelet transform matrix $\tilde{W}$ satisfies the following
condition: there exists a positive constant $\varepsilon$ such that
for any $\alpha\in C$,
\begin{equation}\label{equ4.1}
\|L\circ \tilde{W} \alpha\|_{2}\geq \varepsilon \|\alpha\|_{2},
\end{equation}
where $L$ denotes the local or non-local gradient operator $\nabla$
or $\nabla_{w}$. While $J$ is chosen to be the NL-TV norm, we
further assume that the weight $w(x,y)$ is fixed in this section,
otherwise the function $J$ is non-convex. Then based on some basic
results of convex analysis we can show the existence of the
minimizer. It is stated as follows.

\begin{theorem}\label{thm1}
Assume that the inverse wavelet transform matrix $\tilde{W}$ satisfies the
inequality (\ref{equ4.1}), the minimization problem (\ref{equ3.1})
has at least one solution.
\end{theorem}

\begin{pf}
The problem (\ref{equ3.1}) can be rewritten as
\[
\min_{\alpha\in C} J(f_{0}+\tilde{W}\alpha).
\]
It is simple to verify that the set $C$ is nonempty closed convex.
To prove the existence of the minimizer to problem (\ref{equ3.1}),
it suffices to show that $J(f_{0}+\tilde{W}\alpha)$ is convex and
coercive with respect to $\alpha$. Due to the function $J$ is the TV
or NL-TV regularization term, it satisfies the triangle inequality.
Thus the convexity of $J(f_{0}+\tilde{W}\alpha)$ is a direct result.

Next, we show that $J(f_{0}+\tilde{W}\alpha)$ is coercive. Since
$\tilde{W}$ satisfies the inequality (\ref{equ4.1}),
\[
J(f_{0}+\tilde{W}\alpha)\geq J(\tilde{W}\alpha)-J(-f_{0})\geq \|L\circ \tilde{W} \alpha\|_{2}-J(-f_{0})\geq
\varepsilon \|\alpha\|_{2}-J(f_{0}).
\]
Clearly, if $\|\alpha\|_{1}\rightarrow \infty$, then
$J(f_{0}+\tilde{W}\alpha)\rightarrow \infty$, which completes the
proof.
\end{pf}

The equivalence between the split Bregman algorithm and the
alternating direction method of multipliers (ADMM) has been widely
researched in some previous works \cite{UCLA:ADMM}\cite{MP:DRS}. Then based on the convergence
result of the ADMM proposed by Eckstein and Bertsekas \cite{MP:DRS}, the
convergence of the proposed algorithm can be addressed for the ideal
case where the subproblem in step 2 of Algorithm 1 is solved
exactly. To this end, we review the general theorem by Eckstein and
Bertsekas in which convergence of a generalized version of ADMM is
discussed.

\begin{theorem}\label{thm2}

(Eckstein-Bertsekas) Considering an unconstrained problem of the
form
\begin{equation}\label{equ4.2}
\min_{z\in \mathbb{R}^{n}, u\in \mathbb{R}^{m}}F(z)+G(u), ~~ s.t. ~~
Bz+Du=p,
\end{equation}
where $B\in \mathbb{R}^{l\times n}, D\in \mathbb{R}^{l\times m}$,
and $p\in \mathbb{R}^{l}$. Assume that $F$ and $G$ are closed proper
convex functions, $D$ has full column rank and $F(z)+\|Bz\|_{2}^{2}$
is strictly convex. Suppose $\{\varepsilon_{k}\}$ and $\{\eta_{k}\}$
be two sequences such that $\varepsilon_{k}\geq 0$, $\eta_{k}\geq
0$, $\sum_{k=0}^{\infty}\varepsilon_{k}< \infty$ and
$\sum_{k=0}^{\infty}\eta_{k}< \infty$. Consider three sequences
$\{u^{k}\}, \{z^{k}\}, \{b^{k}\}$ satisfy
\begin{equation}\label{equ4.3}
\left\|u^{k+1}-\textrm{arg}\min_{u}
\left\{G(u)+\frac{\lambda}{2}\|b^{k}+Bz^{k}+Du-p\|^{2}\right\}\right\|\leq
\varepsilon_{k},
\end{equation}
\begin{equation}\label{equ4.4}
\left\|z^{k+1}-\textrm{arg}\min_{z}
\left\{F(z)+\frac{\lambda}{2}\|b^{k}+Bz+Du^{k+1}-p\|^{2}\right\}\right\|\leq
\eta_{k},
\end{equation}
\begin{equation}\label{equ4.5}
b^{k+1}=b^{k} + Bz^{k+1}+Du^{k+1}-p,
\end{equation}
where $\lambda$ is a positive parameter. If the set of the solutions
of (\ref{equ4.2}) is nonempty, the sequence $\{u^{k}\}, \{z^{k}\}$
converges to one solution of (\ref{equ4.2}).
\end{theorem}

Next we discuss the convergence of the proposed algorithm. In order
to guarantee the convergence, we require that the second
sub-minimization problem in (\ref{equ3.6}) is solved exactly enough.
Precisely, let
\begin{equation}\label{equ4.6}
\epsilon_{k}=f^{k}-\textrm{prox}_{\frac{1}{\lambda}
J}(b^{k-1}+f_{0}+\tilde{W}\alpha_{m}^{k}),
\end{equation}
the error should satisfy
\begin{equation}\label{equ4.7}
\Sigma_{k\geq 1}\|\epsilon_{k}\|_{2}<\infty.
\end{equation}
Then the convergence of Algorithm 1 is stated as follows.

\begin{theorem}\label{thm3}
Let the assumptions of Theorem \ref{thm1} hold true, and the
obtained $\{f^{k}\}$ in step 2 of Algorithm 1 satisfies the
formulas (\ref{equ4.6}) and (\ref{equ4.7}). Then the sequence
$\{f^{k}\}$ generated by Algorithm 1 converges to one solution of
(\ref{equ3.1}).
\end{theorem}

\begin{pf}
By identifying $f, \alpha_{m}, -I, \tilde{W}, -f_{0}, J(f),
\iota_{C}(\alpha_{m})$ in problem (\ref{equ3.2}) as $z, u, B, D, p,
F(z), G(u)$ in problem (\ref{equ4.2}), it suffices to show that:

(i) $J(f), \iota_{C}(\alpha_{m})$ are closed proper convex functions
with respect to $f$ and $\alpha_{m}$ respectively;

(ii)the matrix $\tilde{W}$ has full column rank;

(iii)$J(f)+\|f\|^{2}_{2}$ is strictly convex.

$J(f)$ is the TV/NL-TV regularization term. Thus the closed proper
convexity is a direct consequence of the definition. The conditions
in (i) come into existence. The operator $J$ is also a bounded
operator, thus according to the assumption in (\ref{equ4.1}) we
conclude that $\tilde{W}^{T}\tilde{W}>0$ and hence $\tilde{W}$ has
full column rank. Therefore, the conditions in (ii) exist. Since the
function $\|f\|^{2}_{2}$ is strictly convex, the conditions in (iii)
also exist.

Then according to the Eckstein-Bertsekas theorem, the sequence
$\{f^{k}\}$ generated by Algorithm 1 converges to one solution of
(\ref{equ3.1}).

\end{pf}

\section{Numerical examples}\label{sec5}
\setcounter{equation}{0}

In this section, we evaluate the performance of the proposed
algorithm and compare it with the BOS algorithm for wavelet domain
inpainting
\footnote{http://www.math.ucla.edu/~xqzhang/html/code.html} \cite{IPI:NLTVWavelet}. Four
images (size of $256\times 256$): Barbara, Lena, Cameraman and
GoldHill are used for our tests, and all the experiments are
performed under Windows XP and MATLAB 2010 running on a Lenovo
laptop with a Dual Intel Pentium CPU 1.8G and 1 GB of memory.

In all simulated examples presented in the section, we use
Daubechies 7-9 wavelets with symmetric extensions at the boundaries
\cite{CPAM:CD79}, which is adopted in standard JPEG2000 for lossy
compression. 4-scale wavelet decomposition is used for the images in
the test. As is usually done, the standard Peak Signal to Noise
(PSNR) is used to quantify the performance of wavelet coefficient
filling:
\begin{equation}\label{equ5.1}
PSNR(f_{ori},f)=10\lg \left\{\frac{1}{\|f_{ori}-f\|_2}\right\}
\end{equation}
where $f_{ori}$ and $f$ denote the original image and the restored
image respectively.

For the weight function $w(x,y)$ in the NL-TV regularization
(\ref{equ2.7}), we use the same setting as that adopted in \cite{IPI:NLTVWavelet}, i.e.,
the patch size and the searching window for the semi-local weight
are fixed as 5 and 15 and the 10 best neighbors and 4 nearest
neighbors are chosen for the weight computation of each pixel. For
both the BOS algorithm in \cite{IPI:NLTVWavelet} and the proposed algorithm, the TV
denoising (see the formula (\ref{equ2.6}) or step 2 of the proposed
algorithm) is solved by the PDHG method, and the non-local TV
version is solved by the split-Bregman method, for more details
refer to  \cite{IPI:NLTVWavelet}.

In all the following experiments, we choose $\mu=0.05$ for TV,
$\mu=0.01$ for nonlocal TV, and $\delta=1$ for the BOS algorithm
shown in (\ref{equ2.5})-(\ref{equ2.6}). The maximum inner iterations of the
PFBS are set as 10. For the proposed algorithm, the parameter
$\lambda$ for TV is set to be 10, and $[30, 50]$ is a proper
interval for the value of $\lambda$ in the non-local TV case.
Moreover, we use the error limitation as the stopping criterion.
Precisely, $\|P_{I}Wf-\beta\|_{2}<10^{-5}$ and
$\|f-f_{0}-\tilde{W}\alpha_{m}\|_{2}<10^{-5}$ are used for both
algorithms.

In  \cite{IPI:NLTVWavelet}, the author use the image generated only by the known
coefficients as the initial image in the case of some wavelet coefficients loss but
keeping all low-low frequencies. However, the received image is in
very poor quality while part of the LL frequencies are lost.
Therefore, an interpolated image (obtained by applying the nearest
neighbor interpolation on the LL subband) is used as the initial
guess for the BOS algorithm in this case. For our method, we choose
the known component $f_{0}$ by the same strategy.

The Barbara image is used in the first experiment. $\lambda=30$ is adopted
in the proposed algorithm with the non-local regularization. We consider the
case of the lowest HL subband loss for the test image. Due to the
information missing happens in high frequency subands, we observe
that Gibbs artificial or other blur effects appears in the received
image shown in Figure 1(b). The restored images by the TV and NL-TV
methods are shown in the second and third rows of Figure 1. Next, the
case of random loss of wavelet coefficients is considered. Figure
2(a) shows the received image with $60\%$ randomly chosen
coefficients. We observe that many black squares appear in the image
due to the LL frequencies loss. The restored images by using both
approaches are shown in the second and third rows of Figure 2. From
these results we observe that the proposed method produces the
better PSNRs with much less computational time than the BOS
algorithm. In order to make the results more clearly, we also
describe how the PSNR is improved by the proposed method and the BOS
algorithm as the time increases. The plots in Figure 3 show the
evolution of PSNR against CPU time, which verify the efficiency of
the proposed method.

\begin{figure}
  \centering
  \subfigure[Original image]{
    \label{fig:subfig:a} 
    \includegraphics[width=2.0in,clip]{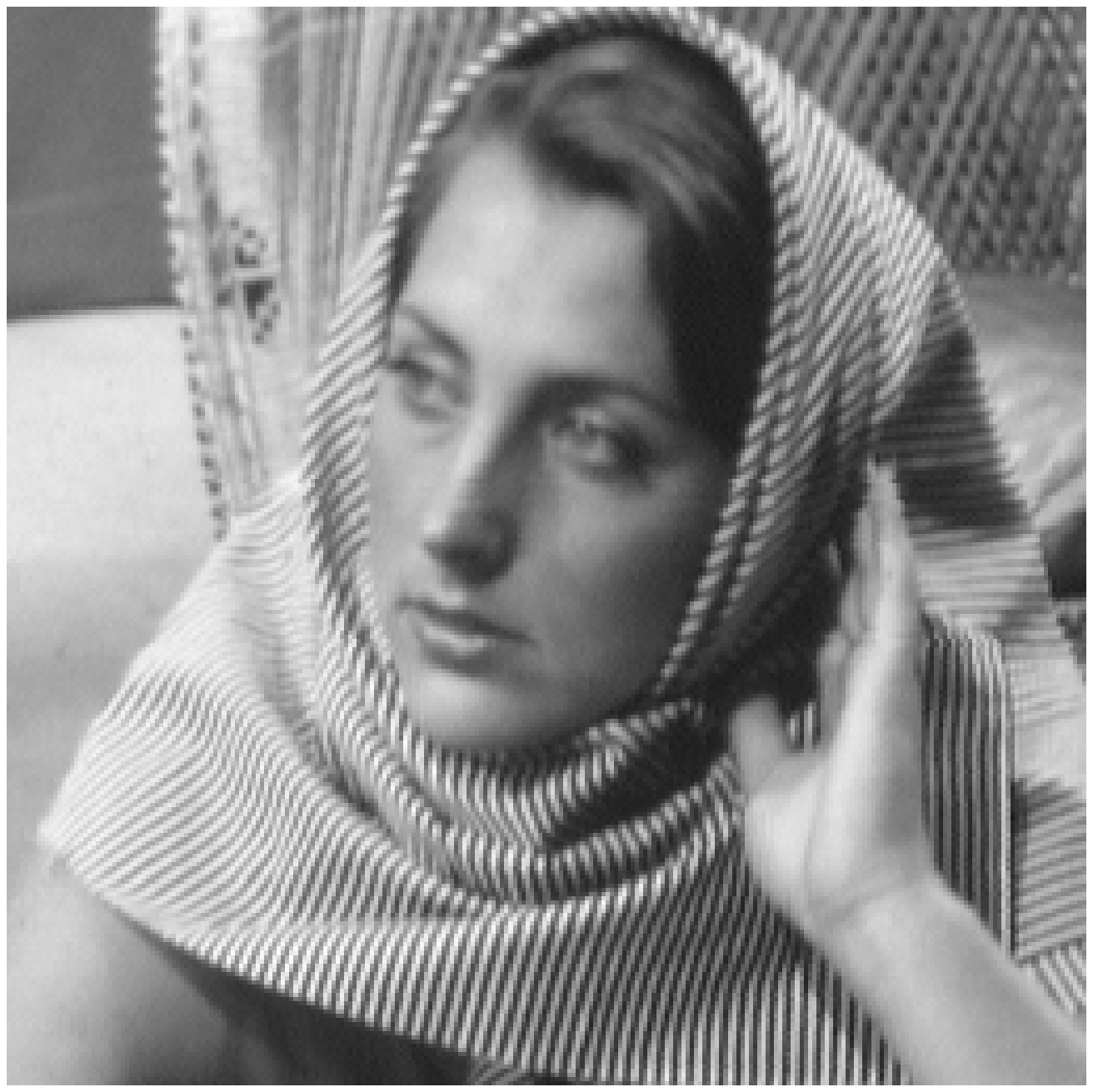}}
  \hspace{0pt}
  \subfigure[Received image, PSNR=29.13dB]{
    \label{fig:subfig:b} 
    \includegraphics[width=2.0in,clip]{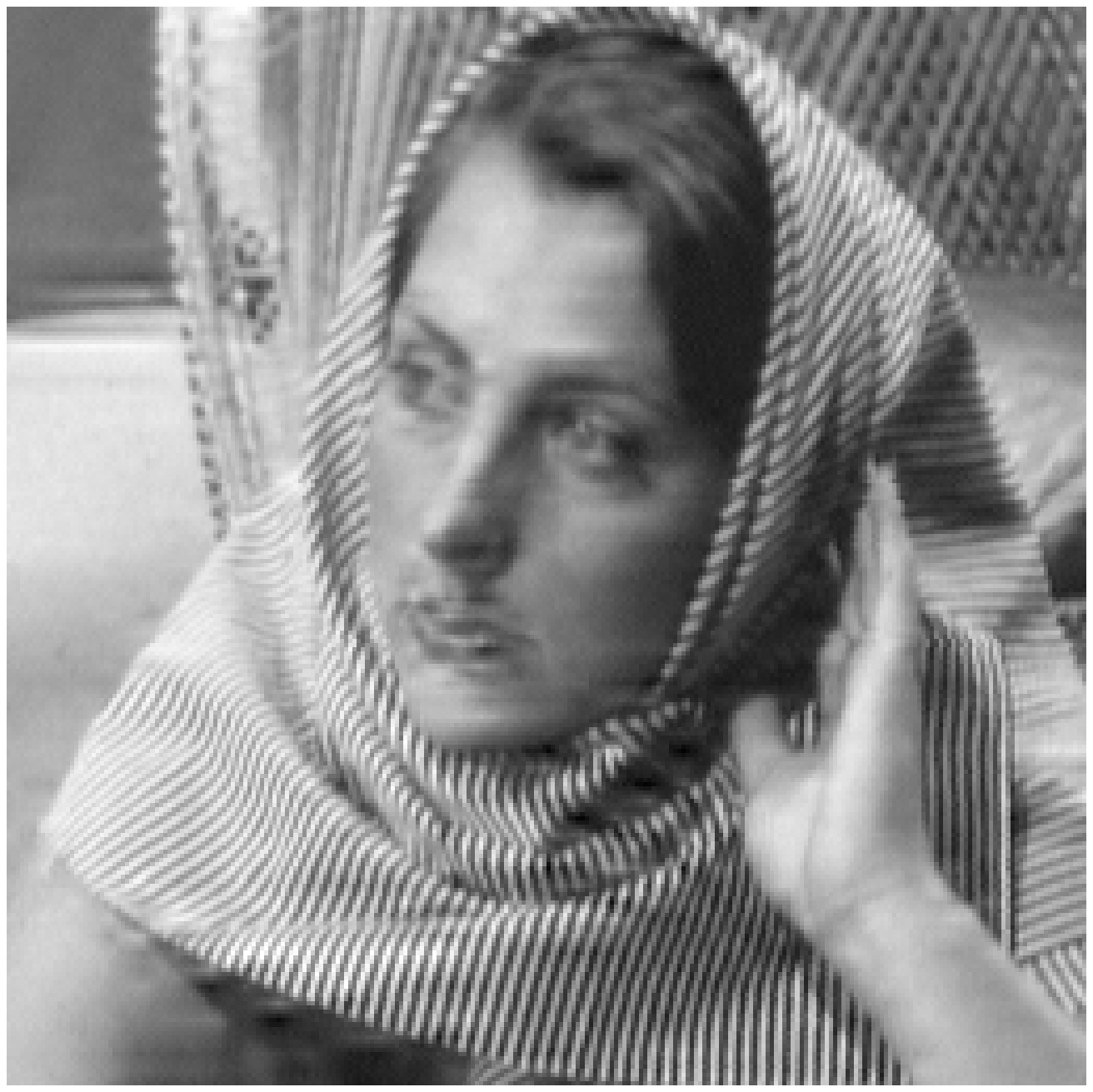}}\\
  \hspace{0pt}
  \subfigure[TV-BOS, PSNR=31.92dB, iter=15, CPU time=128.12s]{
    \label{fig:subfig:c} 
    \includegraphics[width=2.0in,clip]{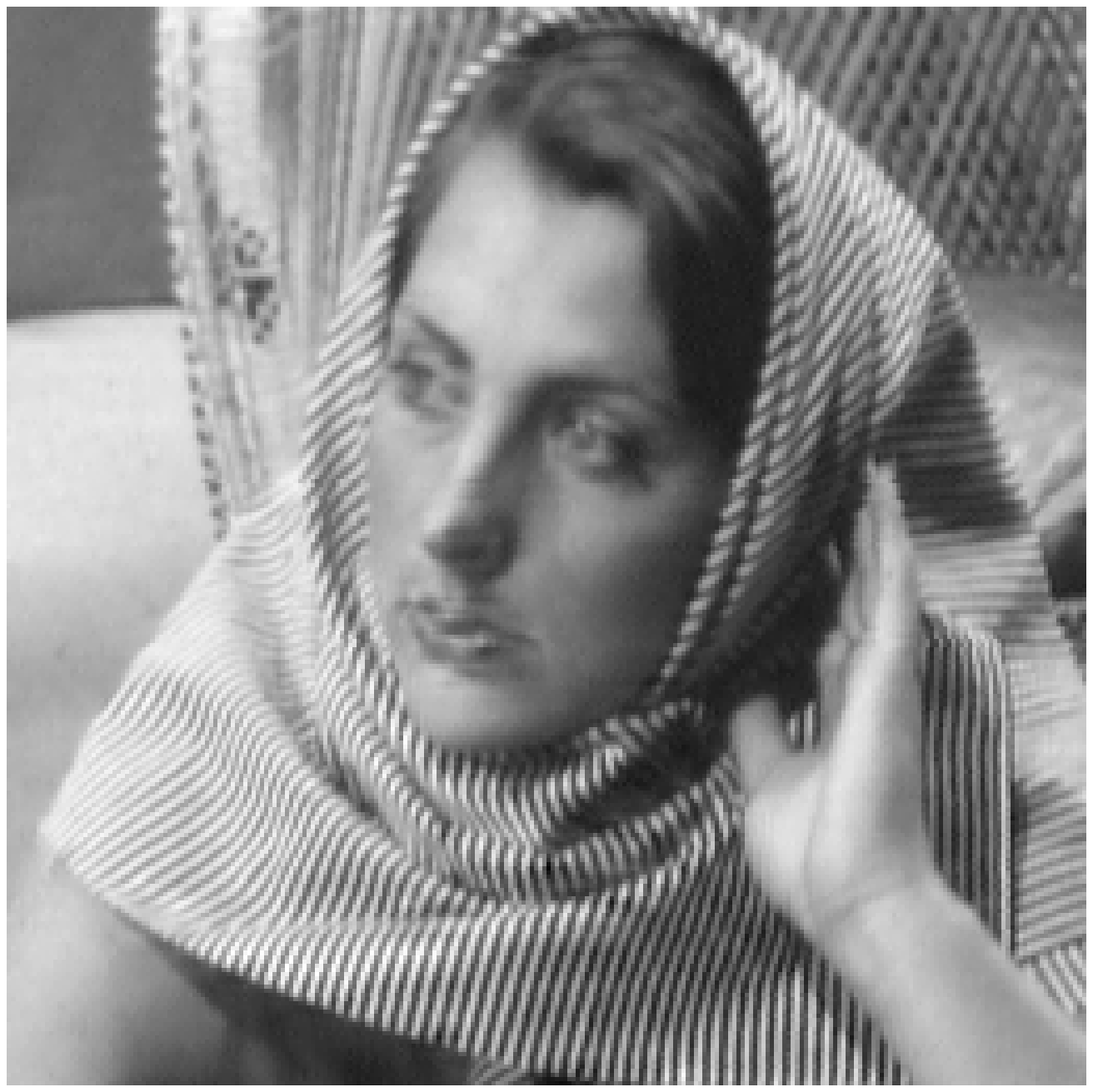}}
  \hspace{0pt}
  \subfigure[TV-Algorithm 1, PSNR=32.03dB, iter=15, CPU time=13.39s]{
    \label{fig:subfig:d} 
    \includegraphics[width=2.0in,clip]{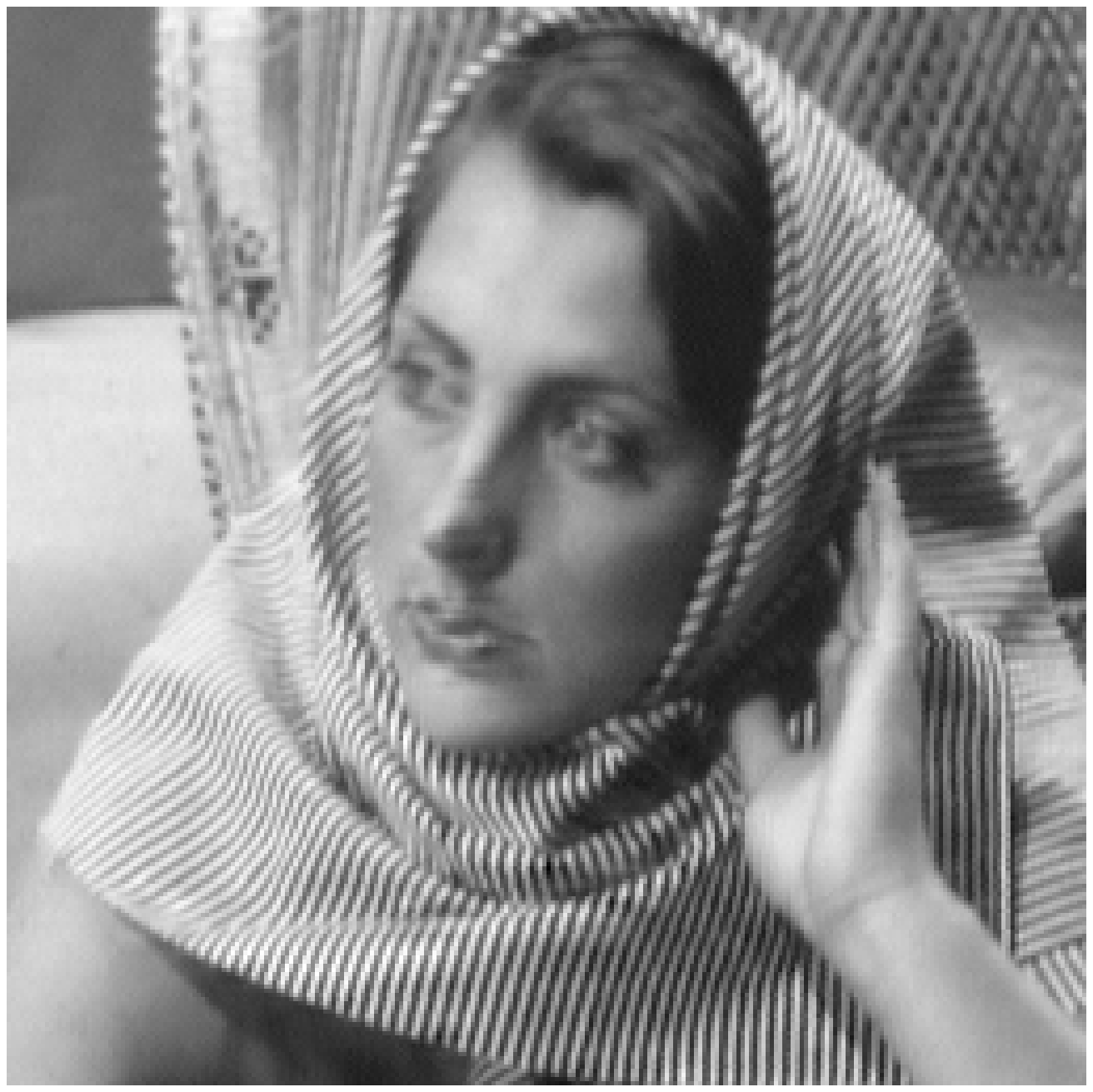}}
  \hspace{0pt}
  \subfigure[NLTV-BOS, PSNR=34.31dB, iter=15, CPU time=778.38s]{
    \label{fig:subfig:e} 
    \includegraphics[width=2.0in,clip]{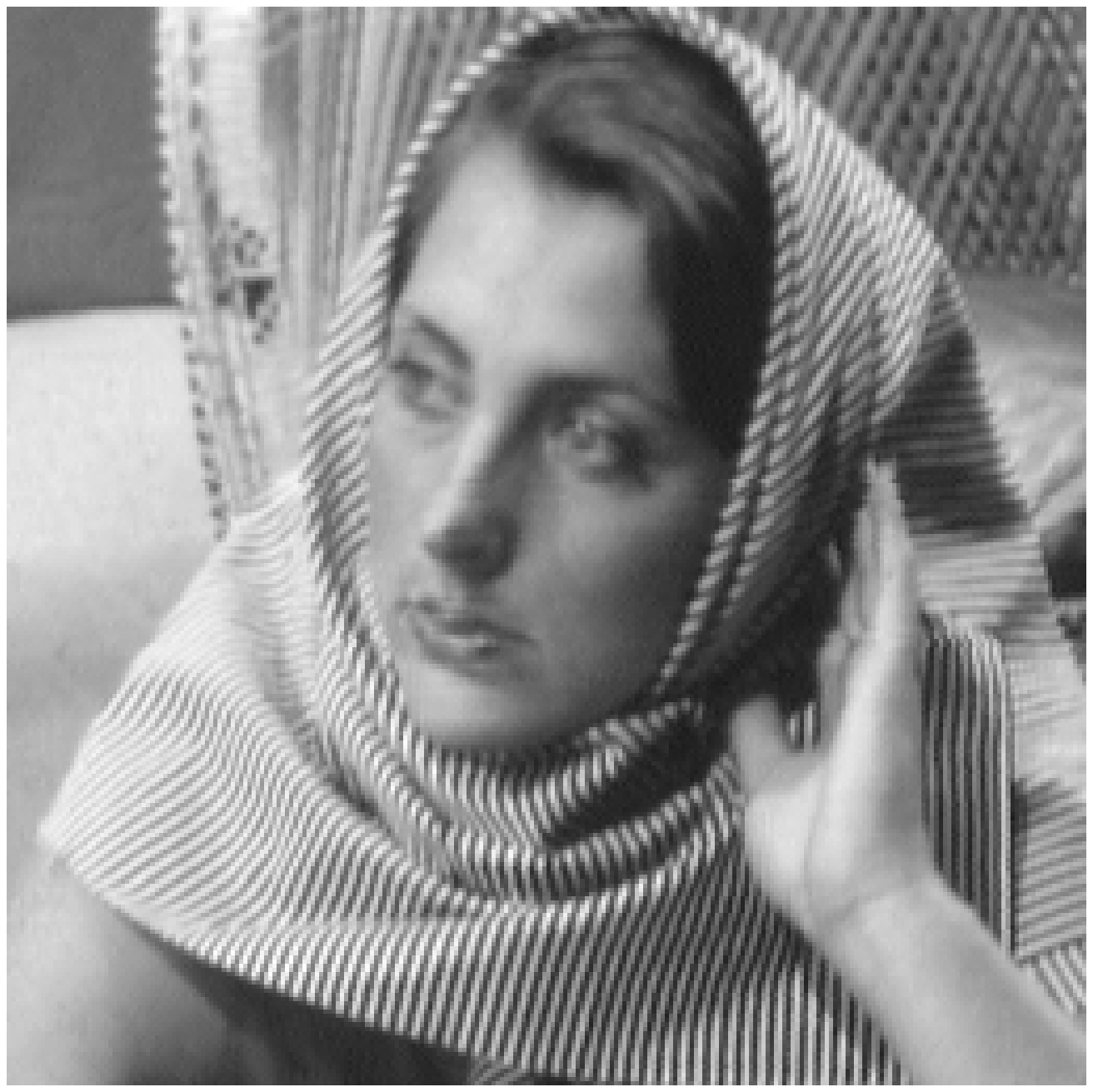}}
  \hspace{0pt}
  \subfigure[NLTV-Algorithm 1, PSNR=34.41dB, iter=25, CPU time=224.73s]{
    \label{fig:subfig:f} 
    \includegraphics[width=2.0in,clip]{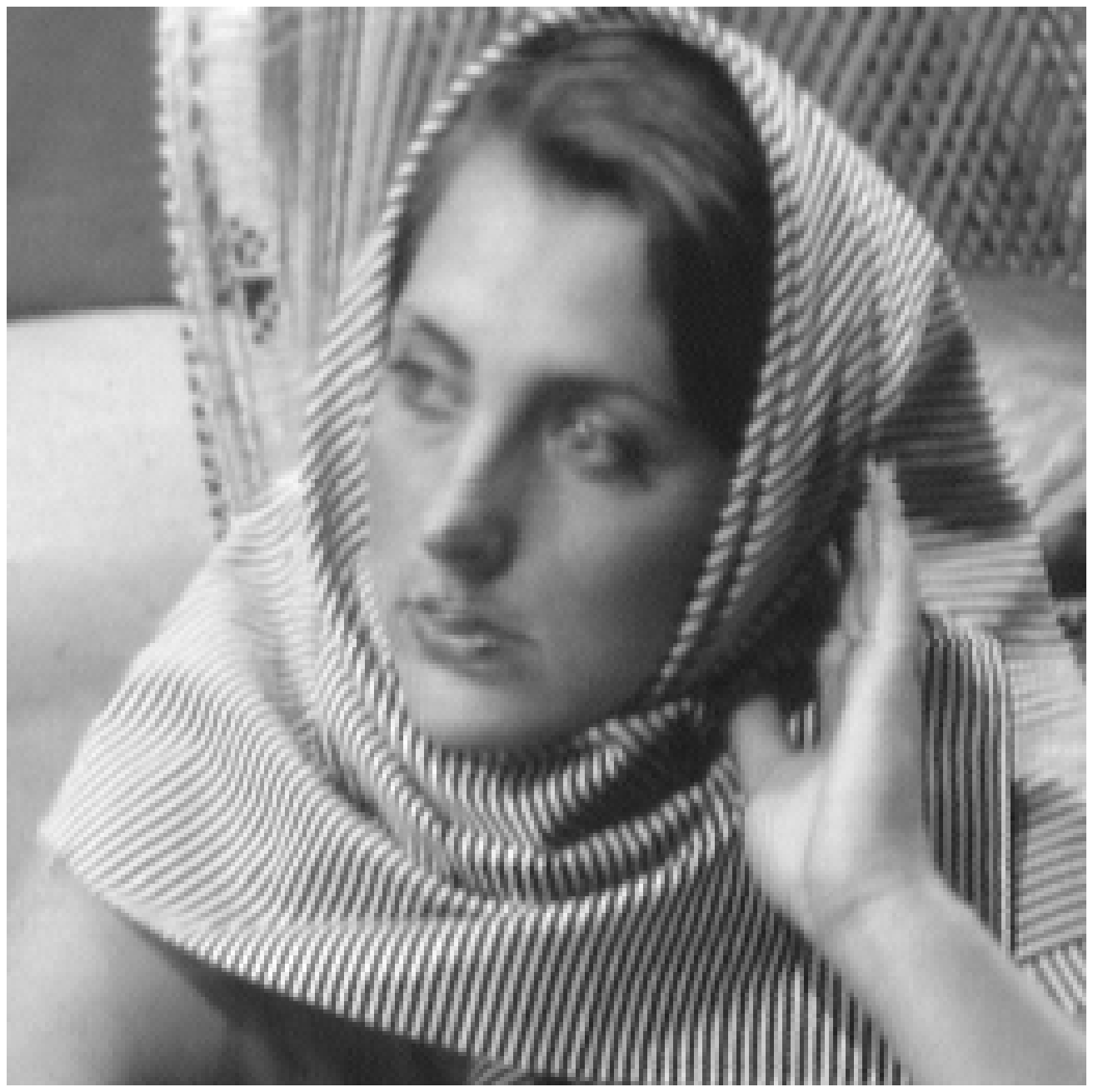}}\\
  \hspace{0pt}
  \label{fig:subfig} 
\caption{Whole HL subband ($32\times 32$) loss. (a)Original image,
(b)received image, (c)the result by BOS algorithm with TV, (d)the
result by Algorithm 1 with TV, (e)the result by BOS algorithm with
NL-TV, (f)the result by Algorithm 1 with NL-TV.}
\end{figure}

\begin{figure}
  \centering
  \subfigure[Received image]{
    \label{fig:subfig:a} 
    \includegraphics[width=2.0in,clip]{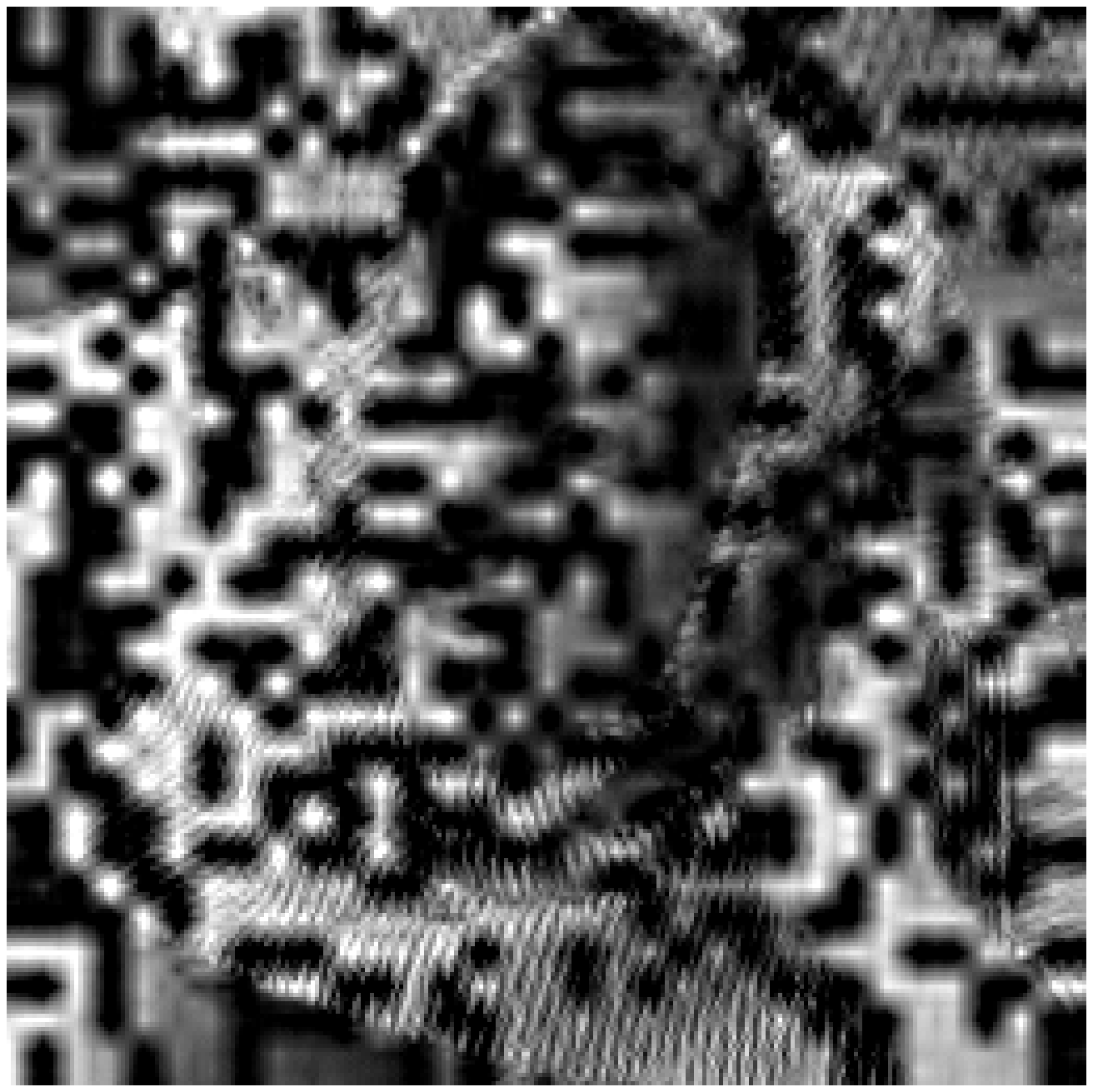}}
  \hspace{0pt}
  \subfigure[Interpolated image, PSNR=19.65dB]{
    \label{fig:subfig:b} 
    \includegraphics[width=2.0in,clip]{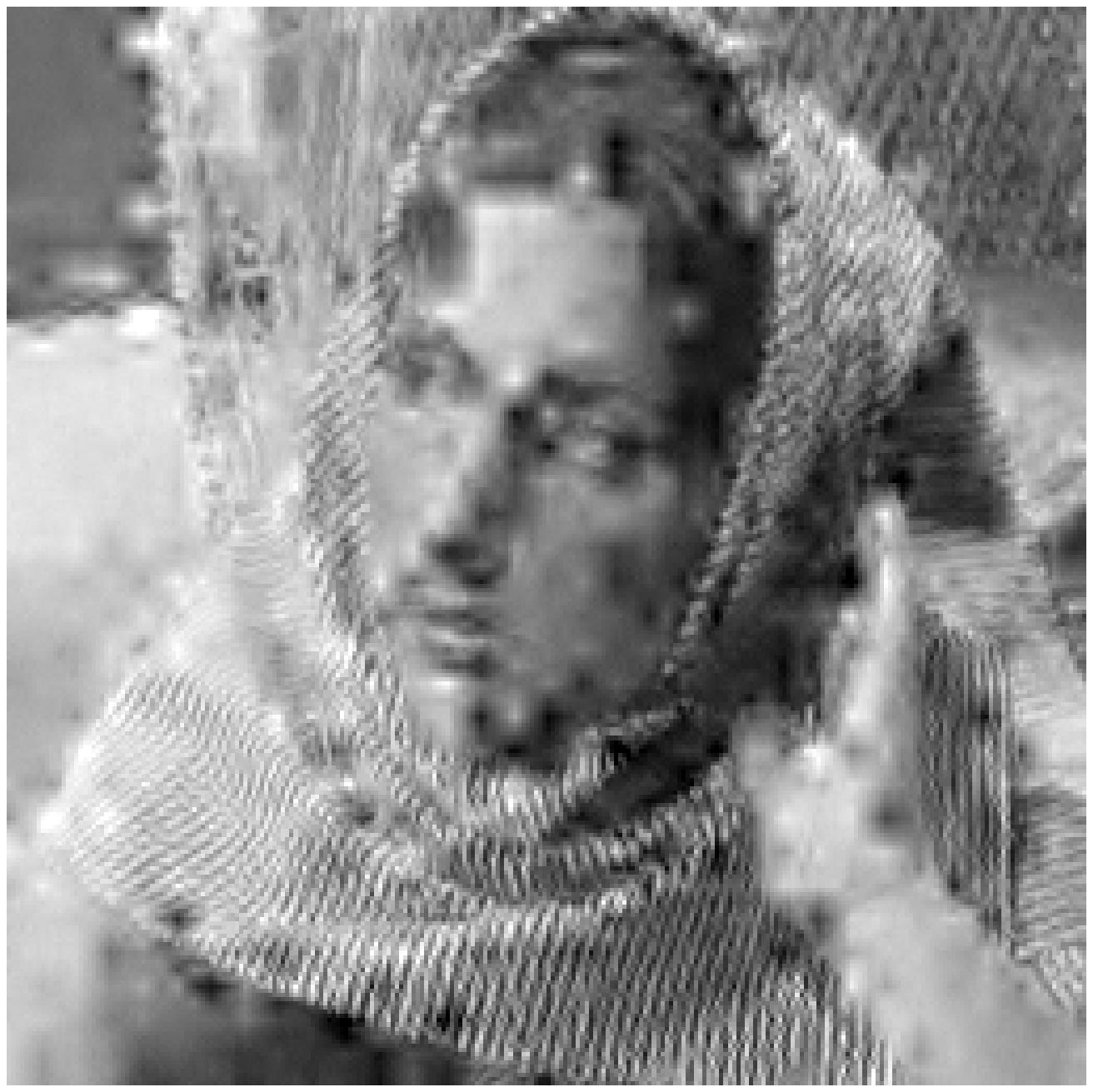}}\\
  \hspace{0pt}
  \subfigure[TV-BOS, PSNR=22.09dB, iter=15, CPU time=131.31s]{
    \label{fig:subfig:c} 
    \includegraphics[width=2.0in,clip]{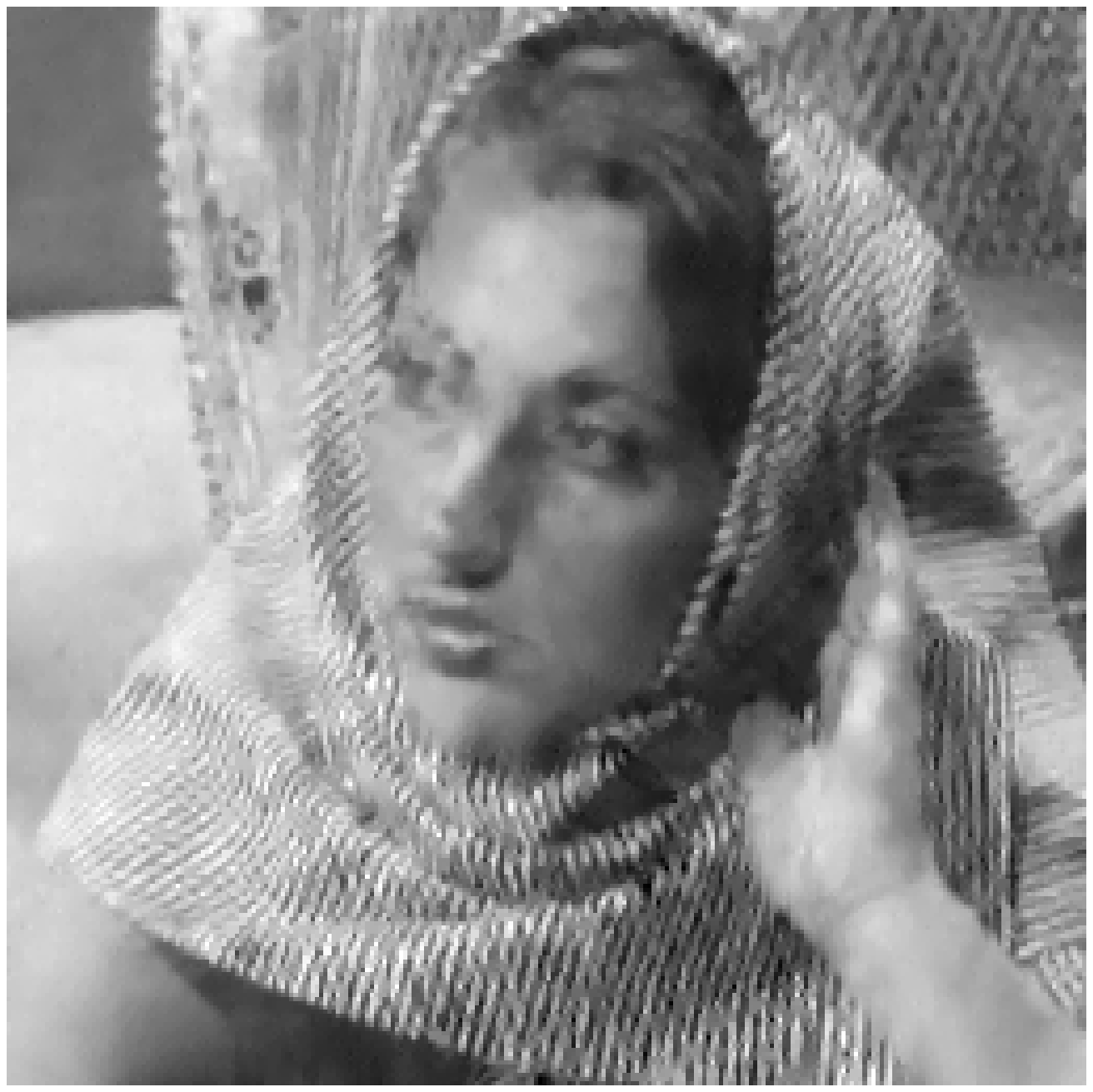}}
  \hspace{0pt}
  \subfigure[TV-Algorithm 1, PSNR=22.15dB, iter=15, CPU time=13.05s]{
    \label{fig:subfig:d} 
    \includegraphics[width=2.0in,clip]{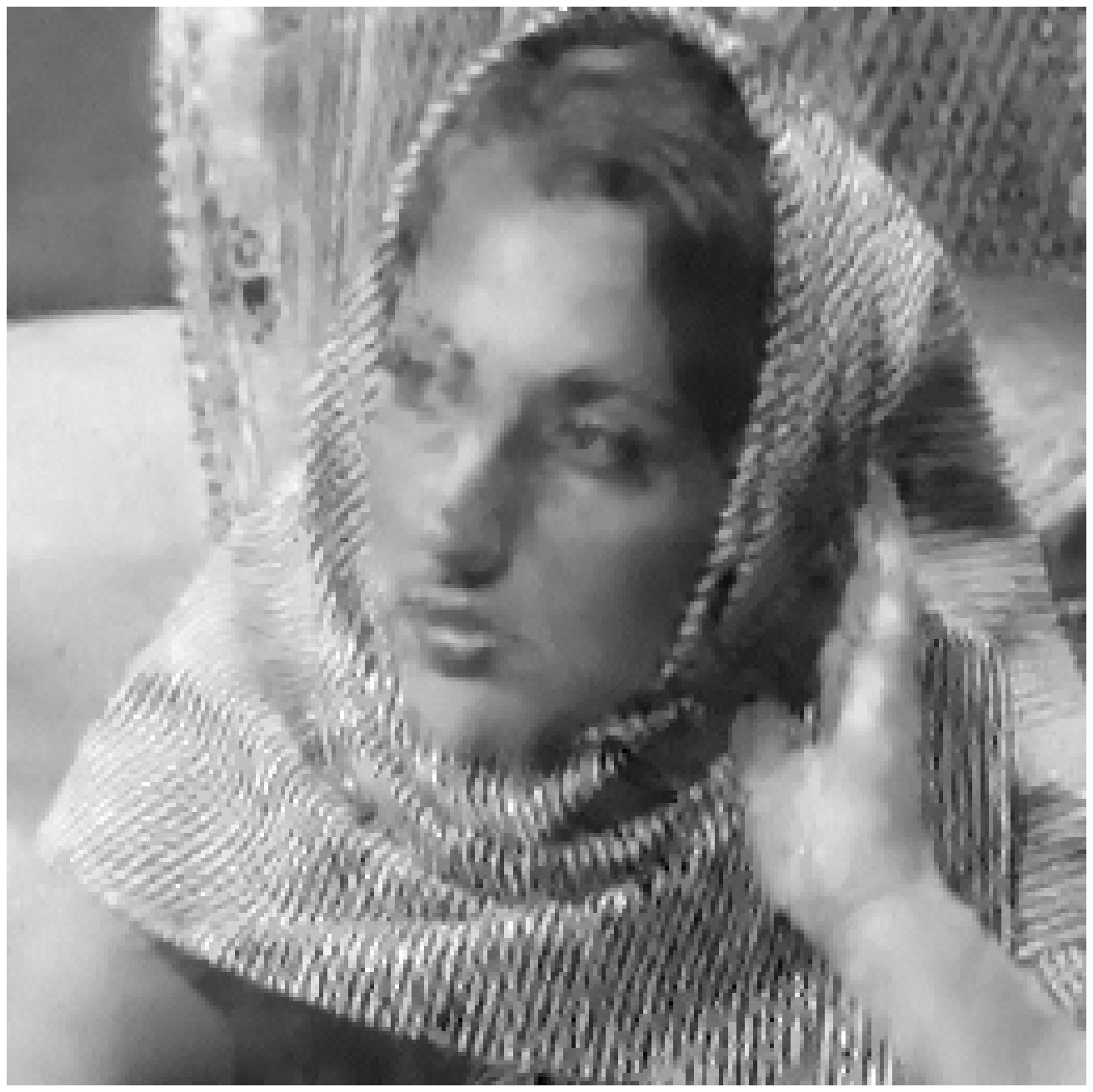}}
  \hspace{0pt}
  \subfigure[NLTV-BOS, PSNR=24.84dB, iter=25, CPU time=1336.7s]{
    \label{fig:subfig:e} 
    \includegraphics[width=2.0in,clip]{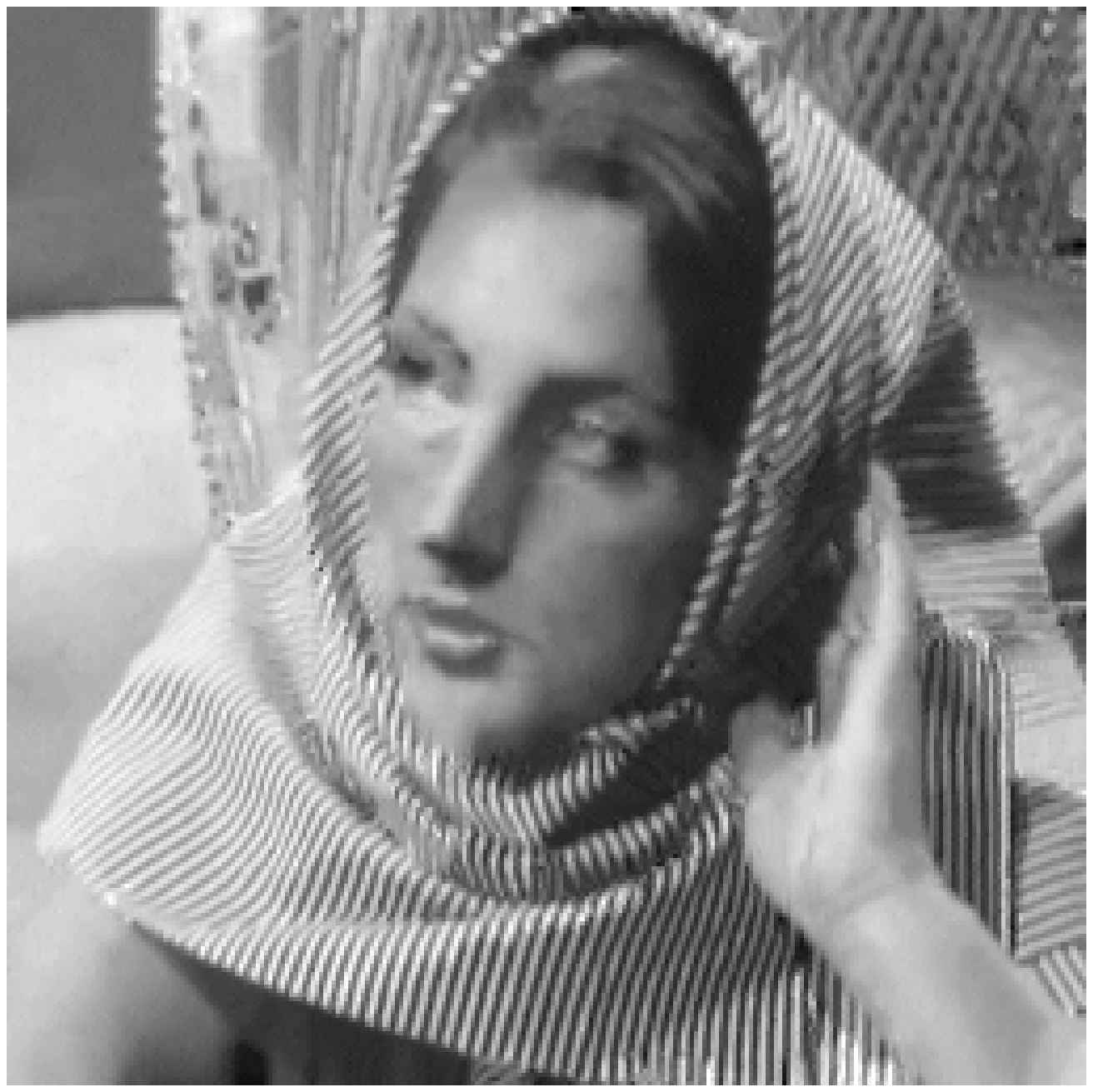}}
  \hspace{0pt}
  \subfigure[NLTV-Algorithm 1, PSNR=24.85dB, iter=25, CPU time=223.61s]{
    \label{fig:subfig:f} 
    \includegraphics[width=2.0in,clip]{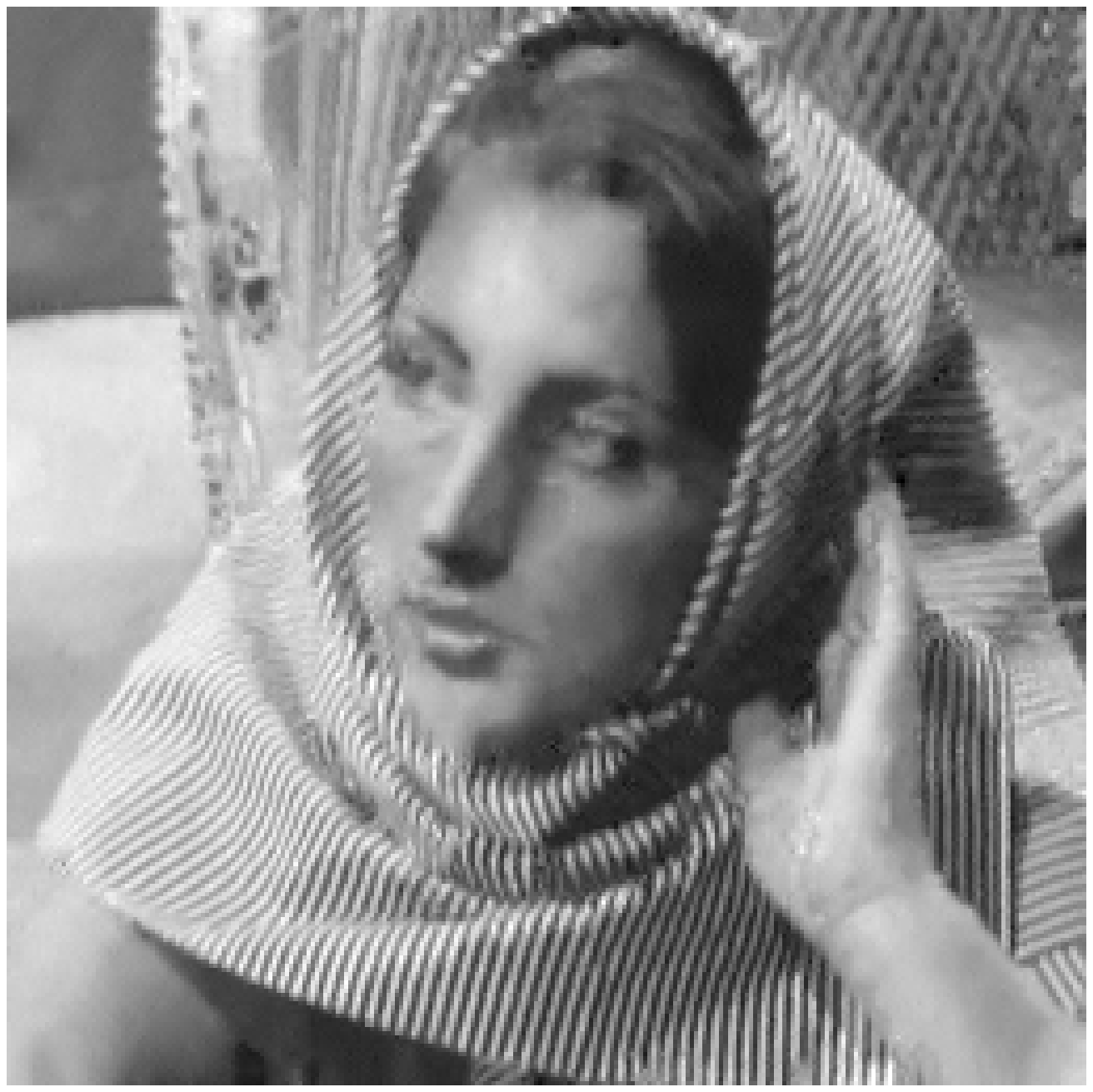}}\\
  \hspace{0pt}
  \label{fig:subfig} 
\caption{Random loss: $60\%$ random chosen frequencies. (a)Received
image, (b)restored image by applying the nearest neighbor
interpolation on the LL subband, (c)the result by BOS algorithm with
TV, (d)the result by Algorithm 1 with TV, (e)the result by BOS
algorithm with NL-TV, (f)the result by Algorithm 1 with NL-TV.}
\end{figure}

\begin{figure}
  \centering
  \subfigure[]{
    \label{fig:subfig:a} 
    \includegraphics[width=2.0in,clip]{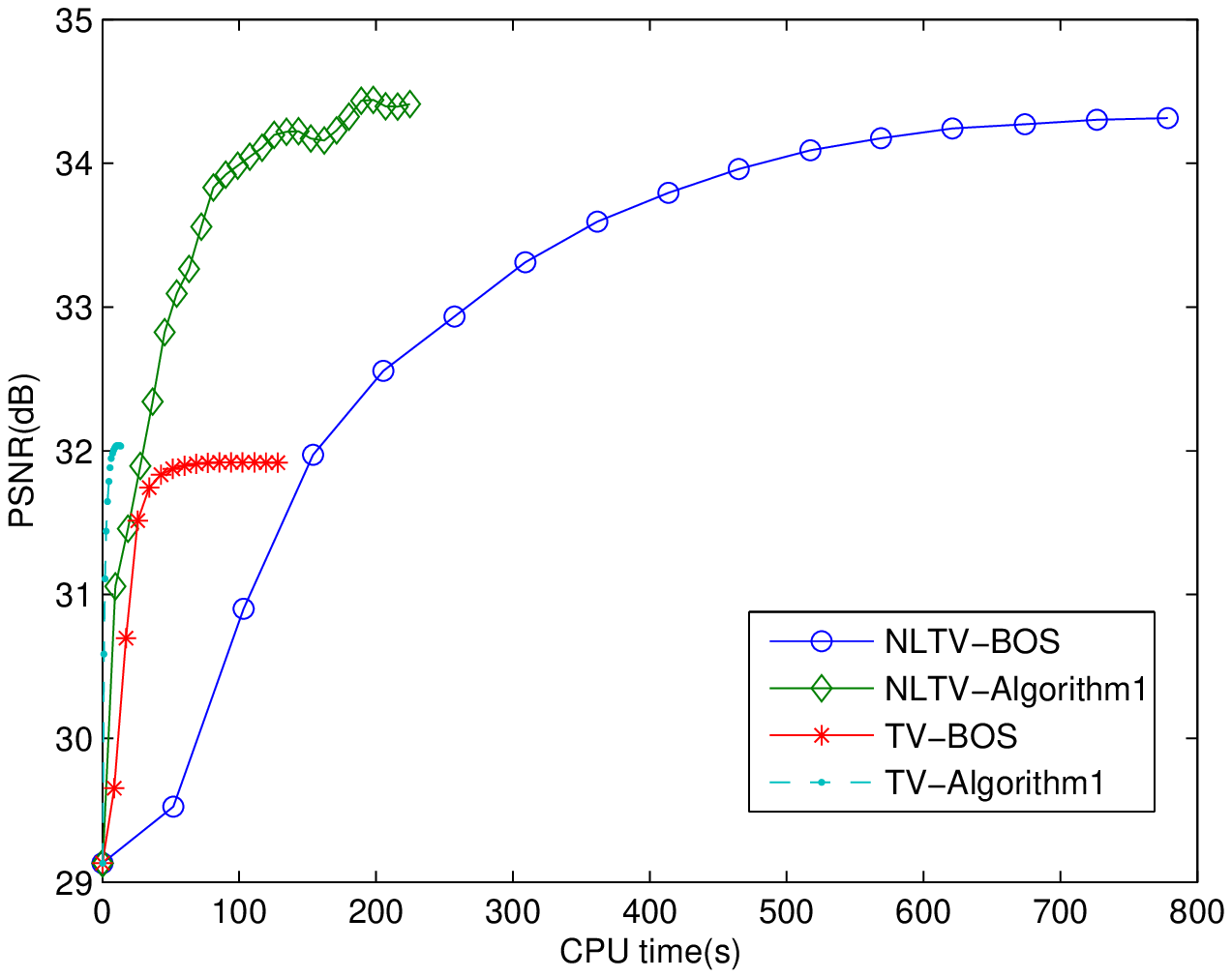}}
  \hspace{0pt}
  \subfigure[]{
    \label{fig:subfig:b} 
    \includegraphics[width=2.0in,clip]{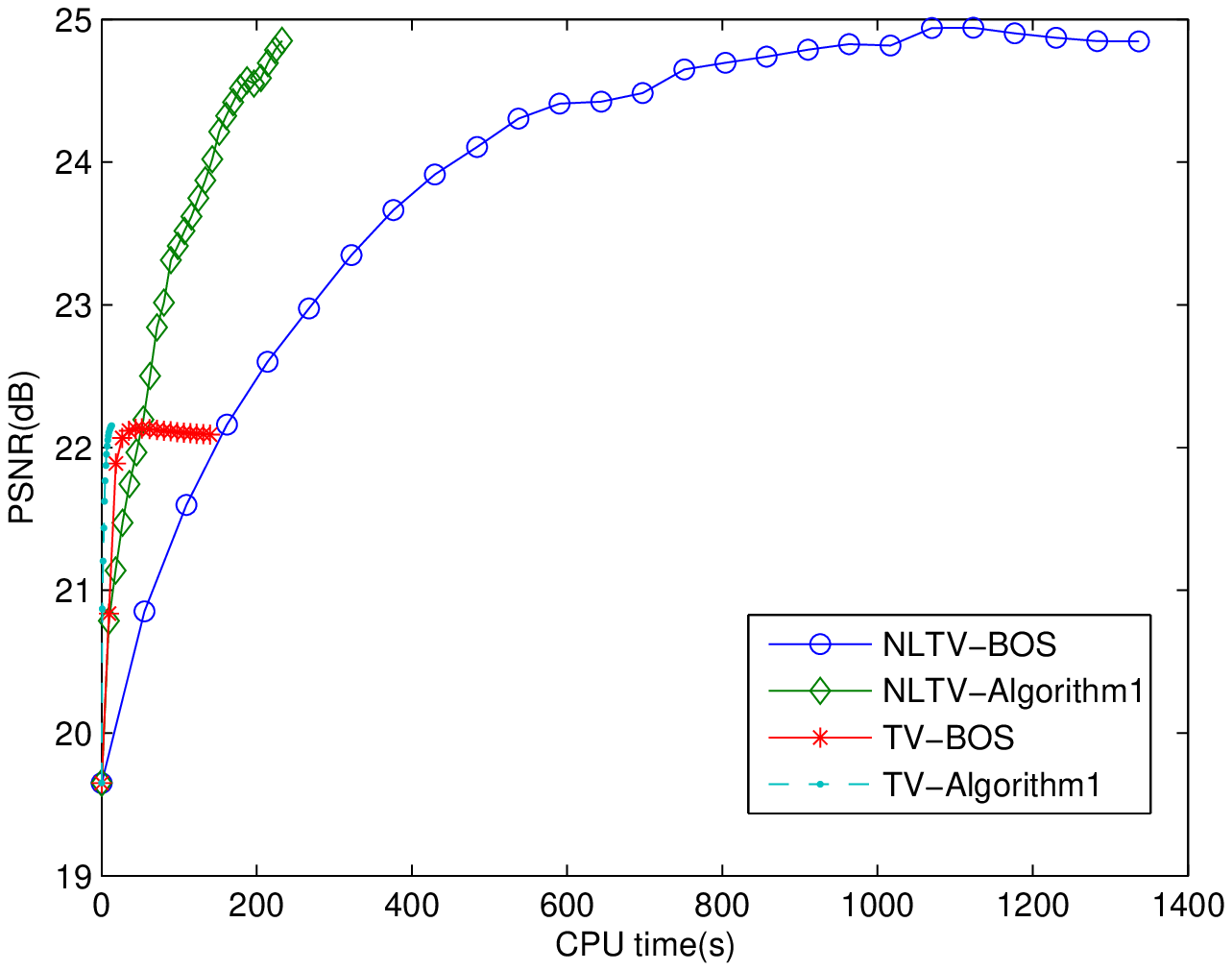}}
  \hspace{0pt}
  \label{fig:subfig} 
\caption{Evolution of PSNR against CPU time with the Barbara image.
(a)The evolution curves of different methods for the received image
in Figure 1, (b)the evolution curves of different methods for the
received image in Figure 2.}
\end{figure}

In the second experiment the standard image Lena is used for our
test. We set $\lambda=50$ for the non-local TV case in the the proposed algorithm.
First we consider the case that the lowest LH subband is lost.
The received image with certain edge blurring effects is shown in Figure 4
(b), and the restored images obtained by the BOS algorithm and our method
are presented in the second and third rows of Figure 4. Second, the case of
random coefficients loss in the high frequencies is considered.
Figure 5(a) shows the $50\%$ randomly chosen high-frequency coefficients and all low-low frequencies
on the top left, and 5(b) is the corresponding image. Both algorithms with TV/NL-TV regularization
are implemented and the results are shown as Figure 5(c)-(f). The plots in Figure 6
show the evolution of PSNR against CPU time. Once again we observe that the proposed method
can achieve the same PSNRs much faster than the BOS algorithm.

\begin{figure}
  \centering
  \subfigure[Original image]{
    \label{fig:subfig:a} 
    \includegraphics[width=2.0in,clip]{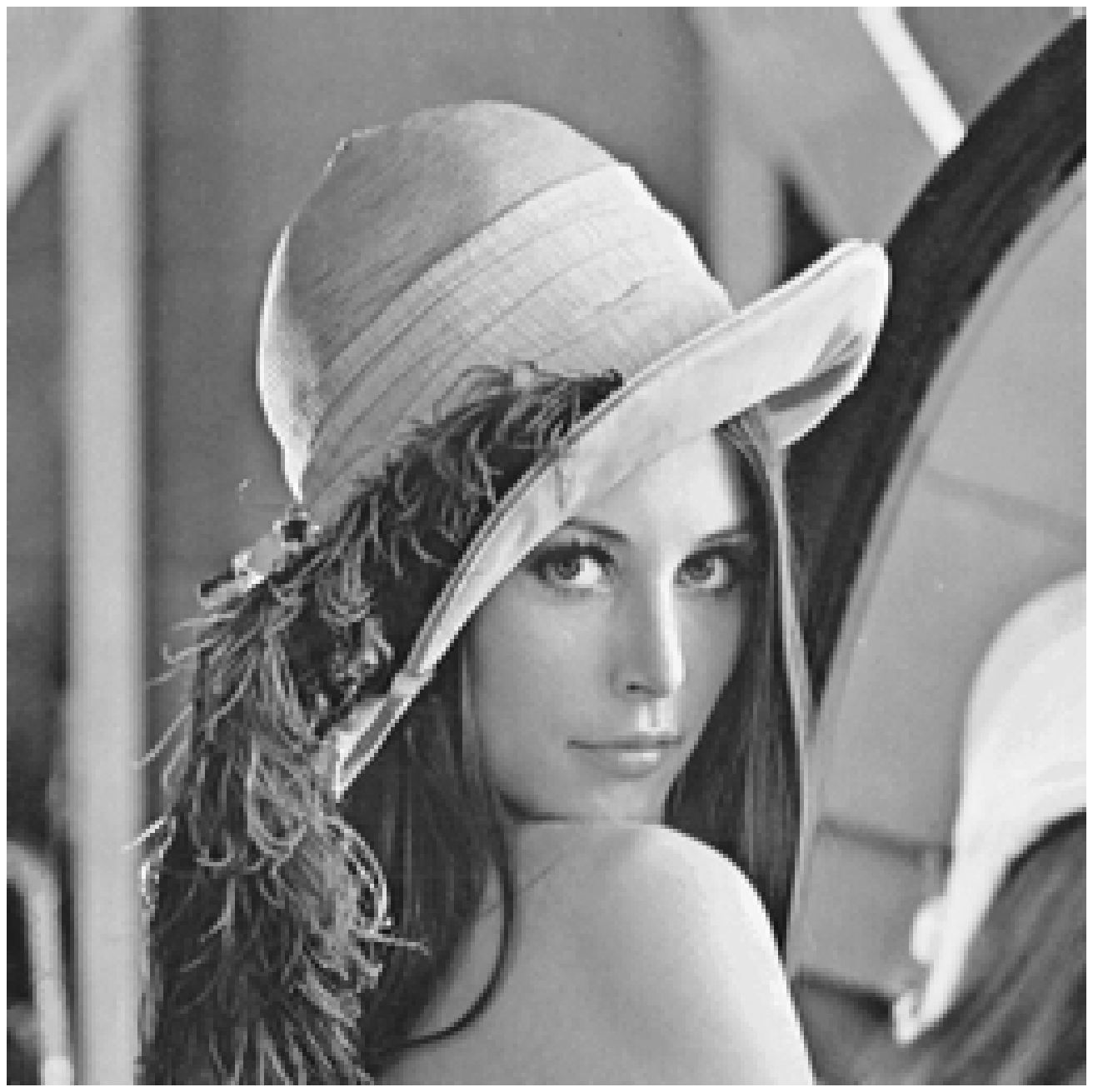}}
  \hspace{0pt}
  \subfigure[Received image, PSNR=24.88dB]{
    \label{fig:subfig:b} 
    \includegraphics[width=2.0in,clip]{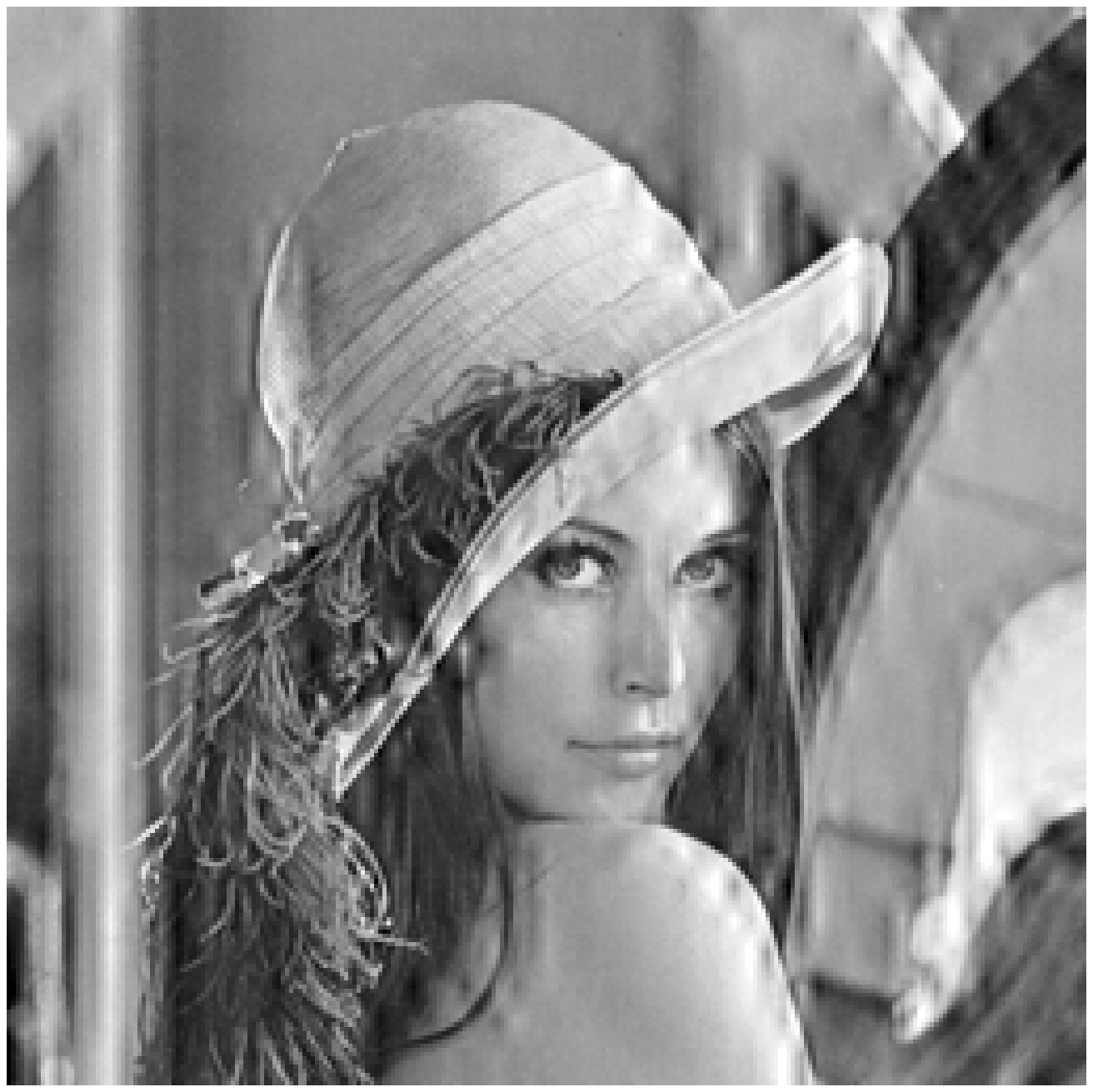}}\\
  \hspace{0pt}
  \subfigure[TV-BOS, PSNR=31.34dB, iter=15, CPU time=129.08s]{
    \label{fig:subfig:c} 
    \includegraphics[width=2.0in,clip]{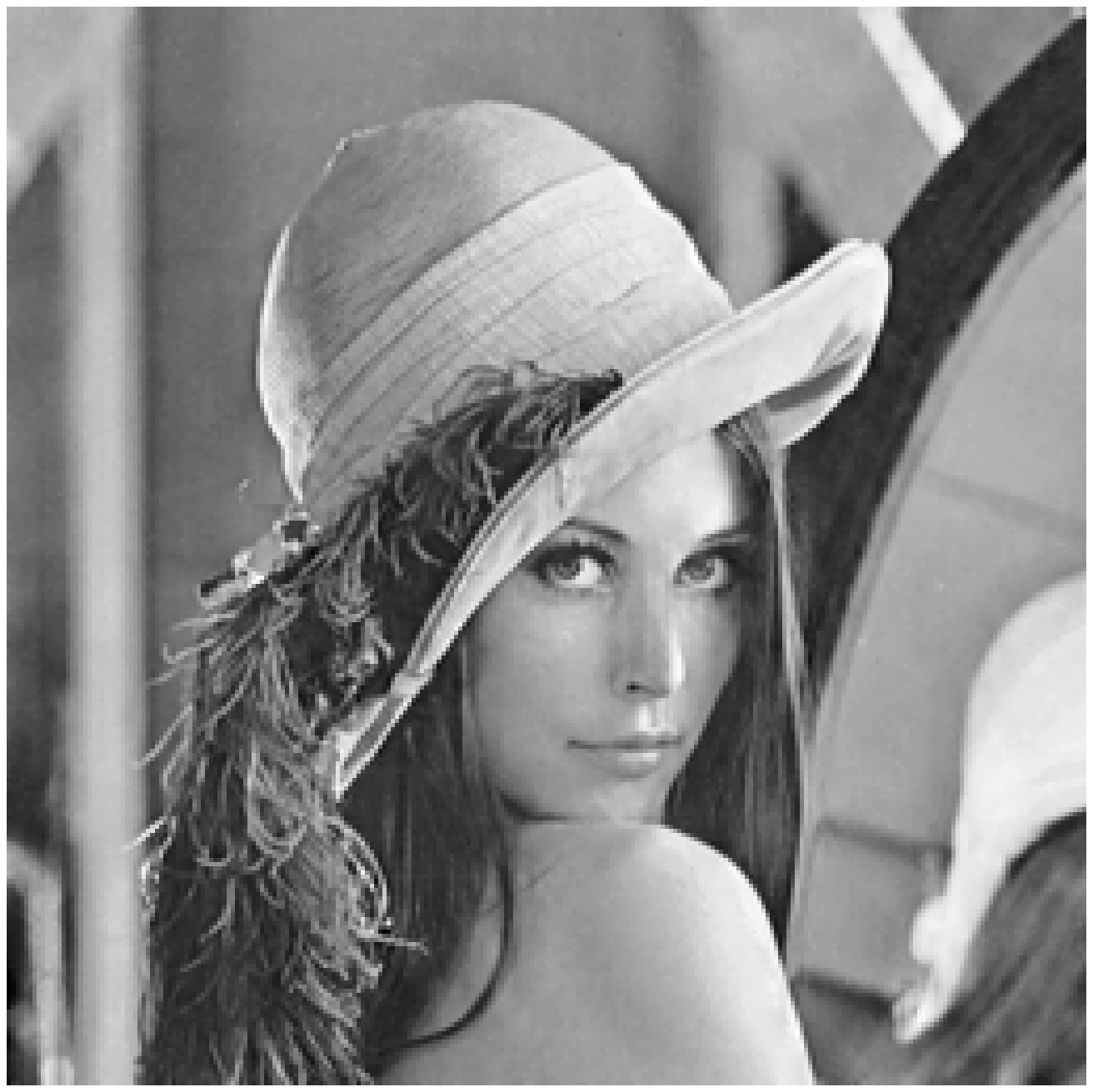}}
  \hspace{0pt}
  \subfigure[TV-Algorithm 1, PSNR=31.32dB, iter=20, CPU time=19.37s]{
    \label{fig:subfig:d} 
    \includegraphics[width=2.0in,clip]{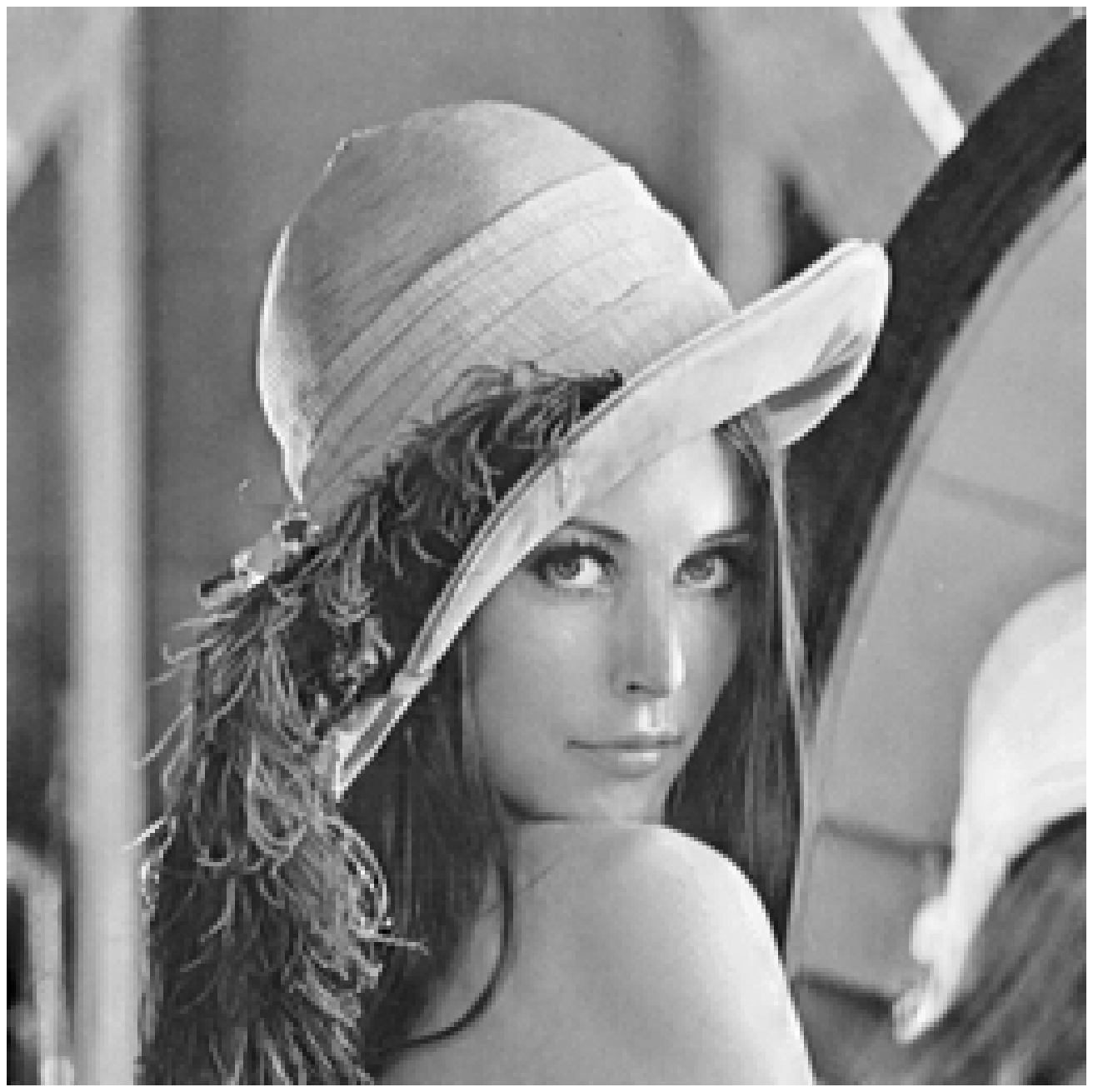}}
  \hspace{0pt}
  \subfigure[NLTV-BOS, PSNR=33.36dB, iter=15, CPU time=835.16s]{
    \label{fig:subfig:e} 
    \includegraphics[width=2.0in,clip]{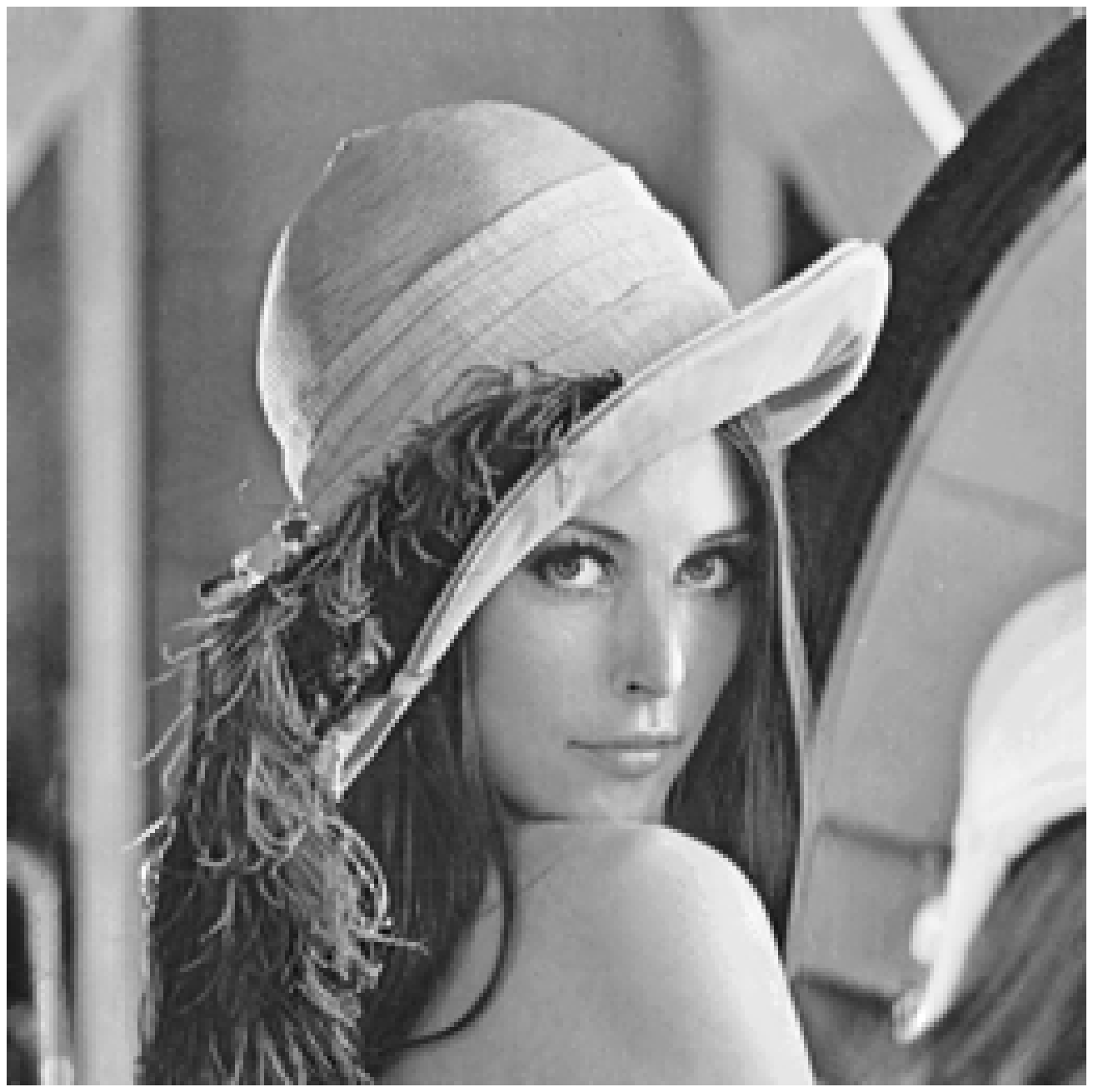}}
  \hspace{0pt}
  \subfigure[NLTV-Algorithm 1, PSNR=33.39dB, iter=25, CPU time=222.42s]{
    \label{fig:subfig:f} 
    \includegraphics[width=2.0in,clip]{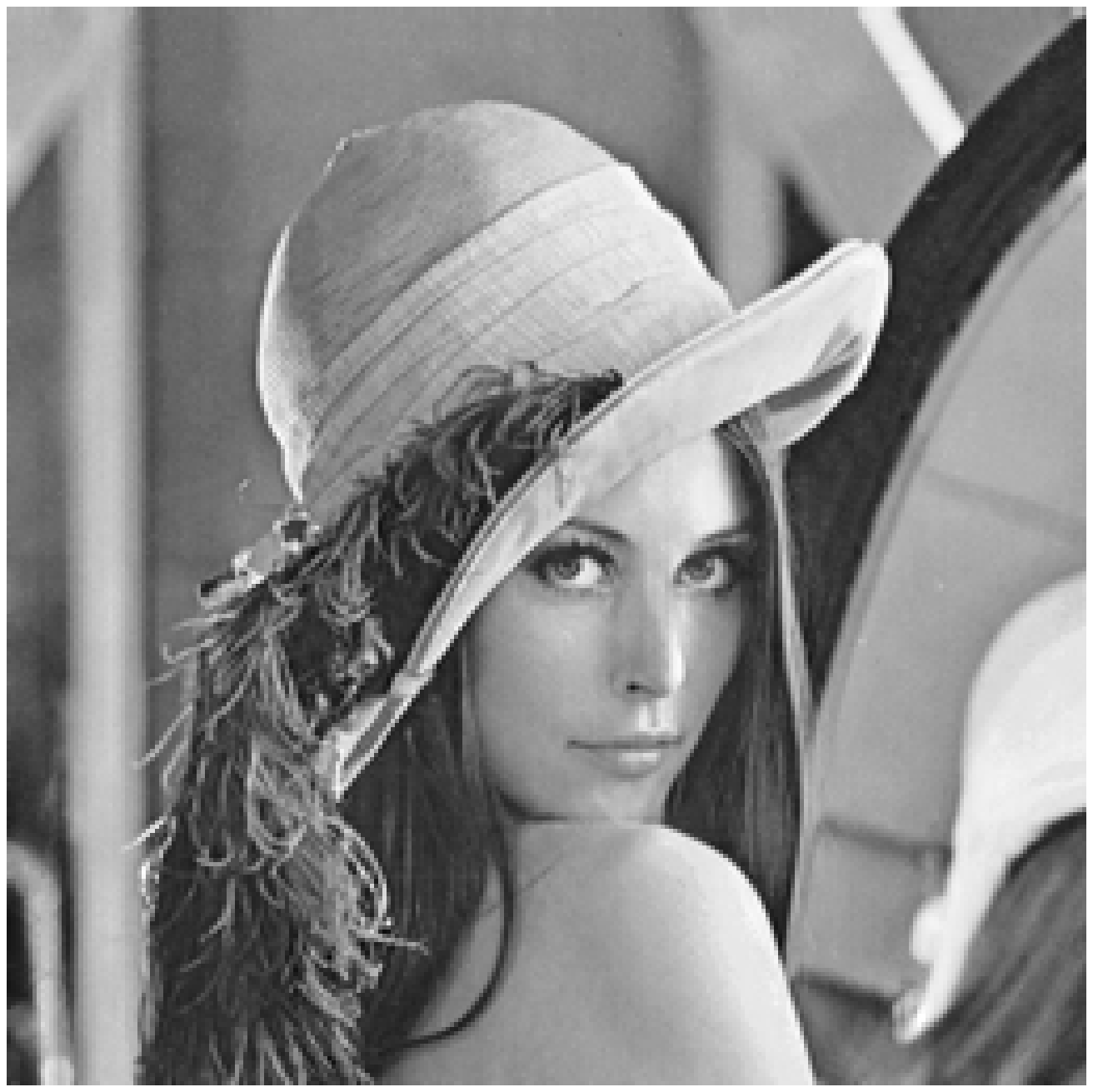}}\\
  \hspace{0pt}
  \label{fig:subfig} 
\caption{Whole LH subband ($32\times 32$) loss. (a)Original image,
(b)received image, (c)the result by BOS algorithm with TV, (d)the
result by Algorithm 1 with TV, (e)the result by BOS algorithm with
NL-TV, (f)the result by Algorithm 1 with NL-TV.}
\end{figure}

\begin{figure}
  \centering
  \subfigure[Received wavelet coefficients]{
    \label{fig:subfig:a} 
    \includegraphics[width=2.0in,clip]{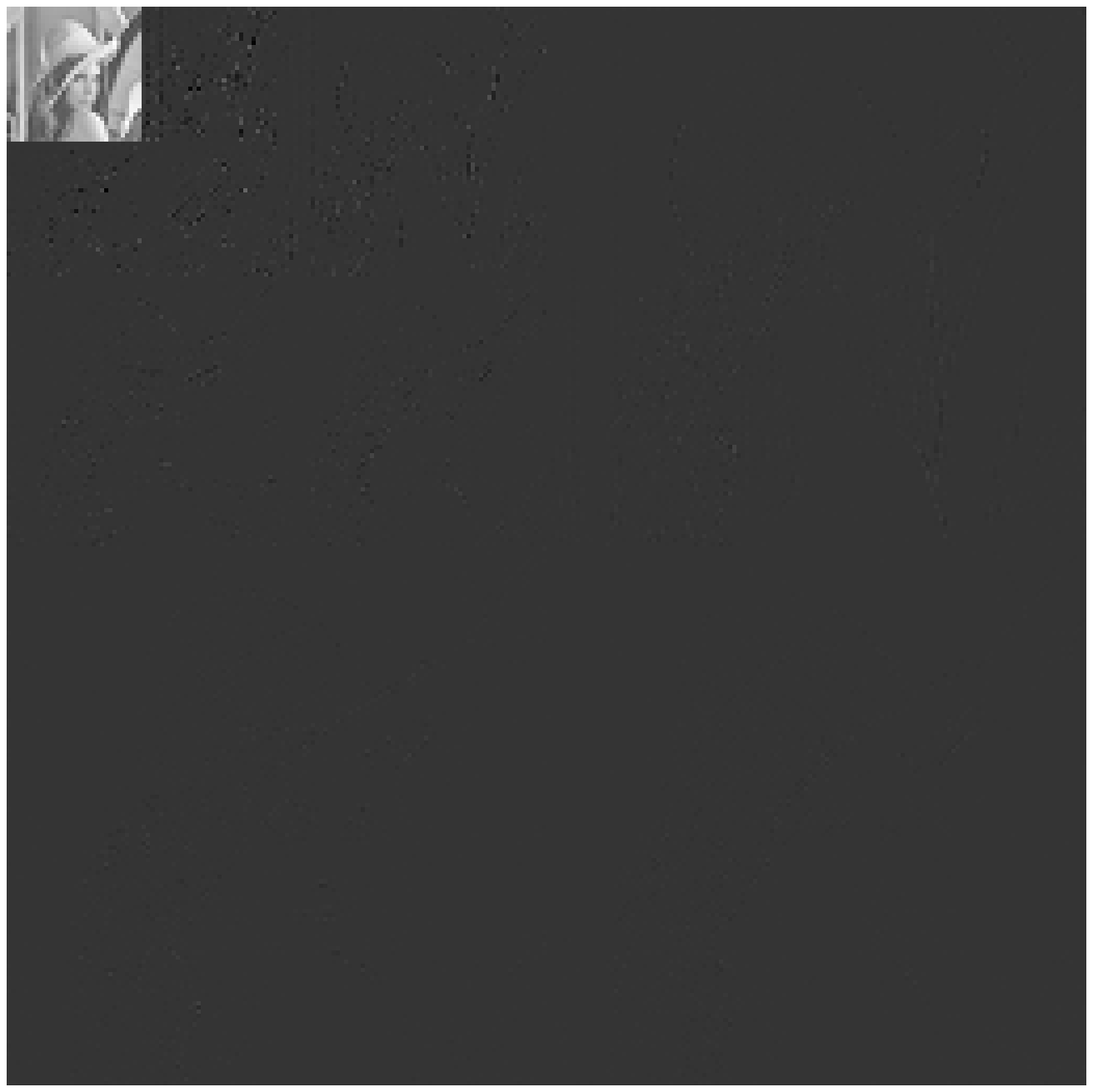}}
  \hspace{0pt}
  \subfigure[Received image, PSNR=22.65dB]{
    \label{fig:subfig:b} 
    \includegraphics[width=2.0in,clip]{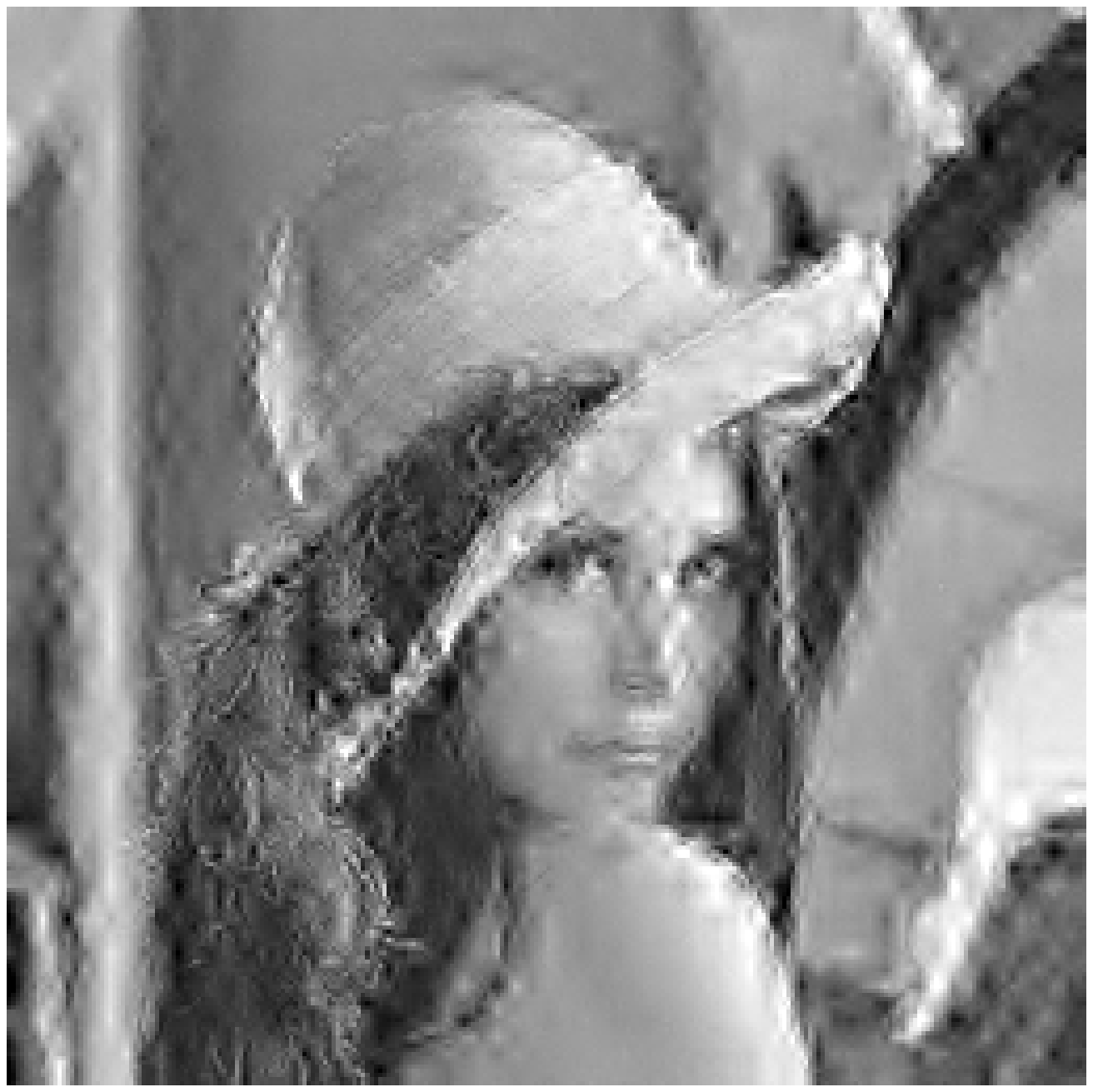}}\\
  \hspace{0pt}
  \subfigure[TV-BOS, PSNR=26.10dB, iter=15, CPU time=130.33s]{
    \label{fig:subfig:c} 
    \includegraphics[width=2.0in,clip]{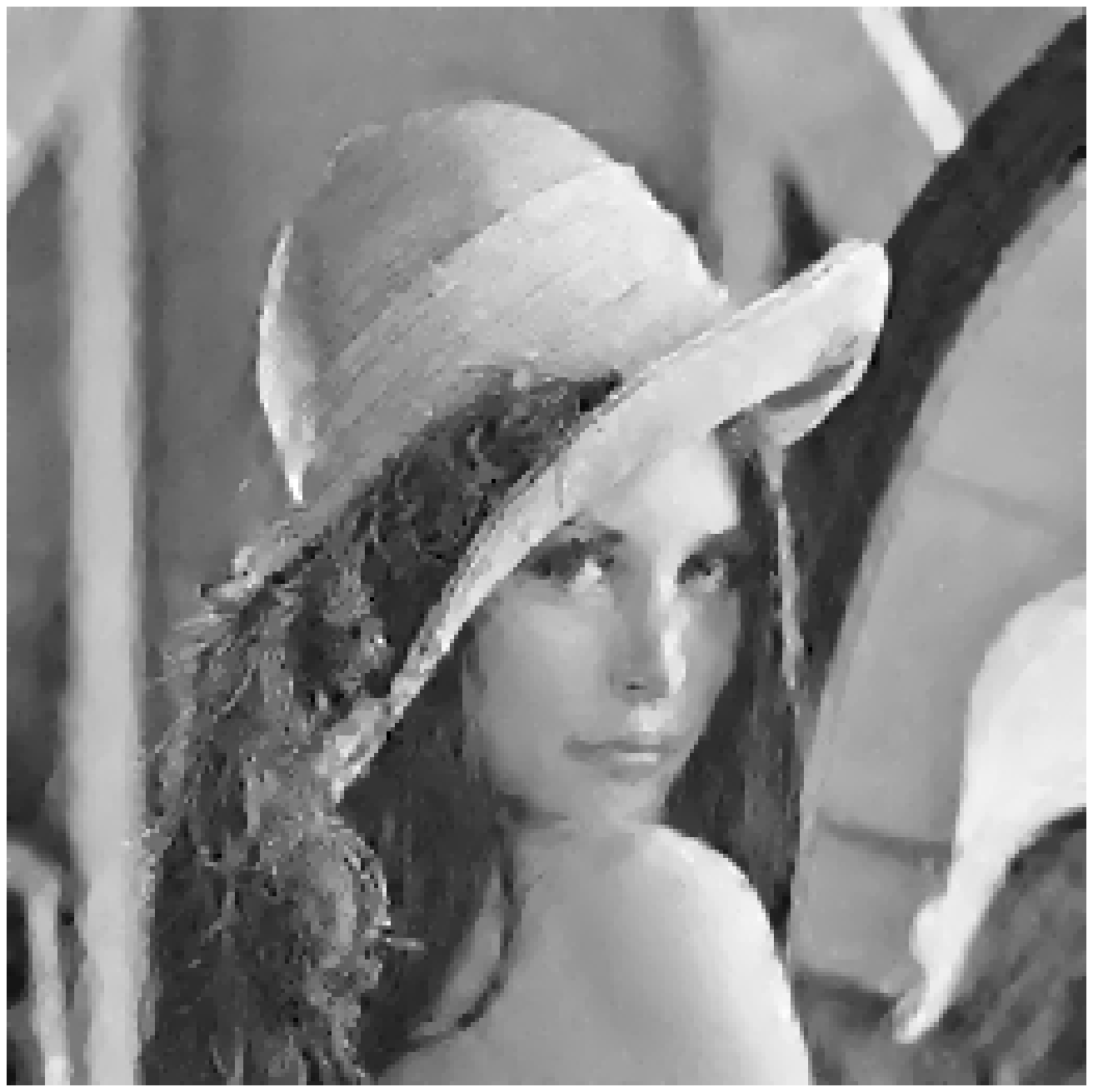}}
  \hspace{0pt}
  \subfigure[TV-Algorithm 1, PSNR=26.08dB, iter=15, CPU time=13.61s]{
    \label{fig:subfig:d} 
    \includegraphics[width=2.0in,clip]{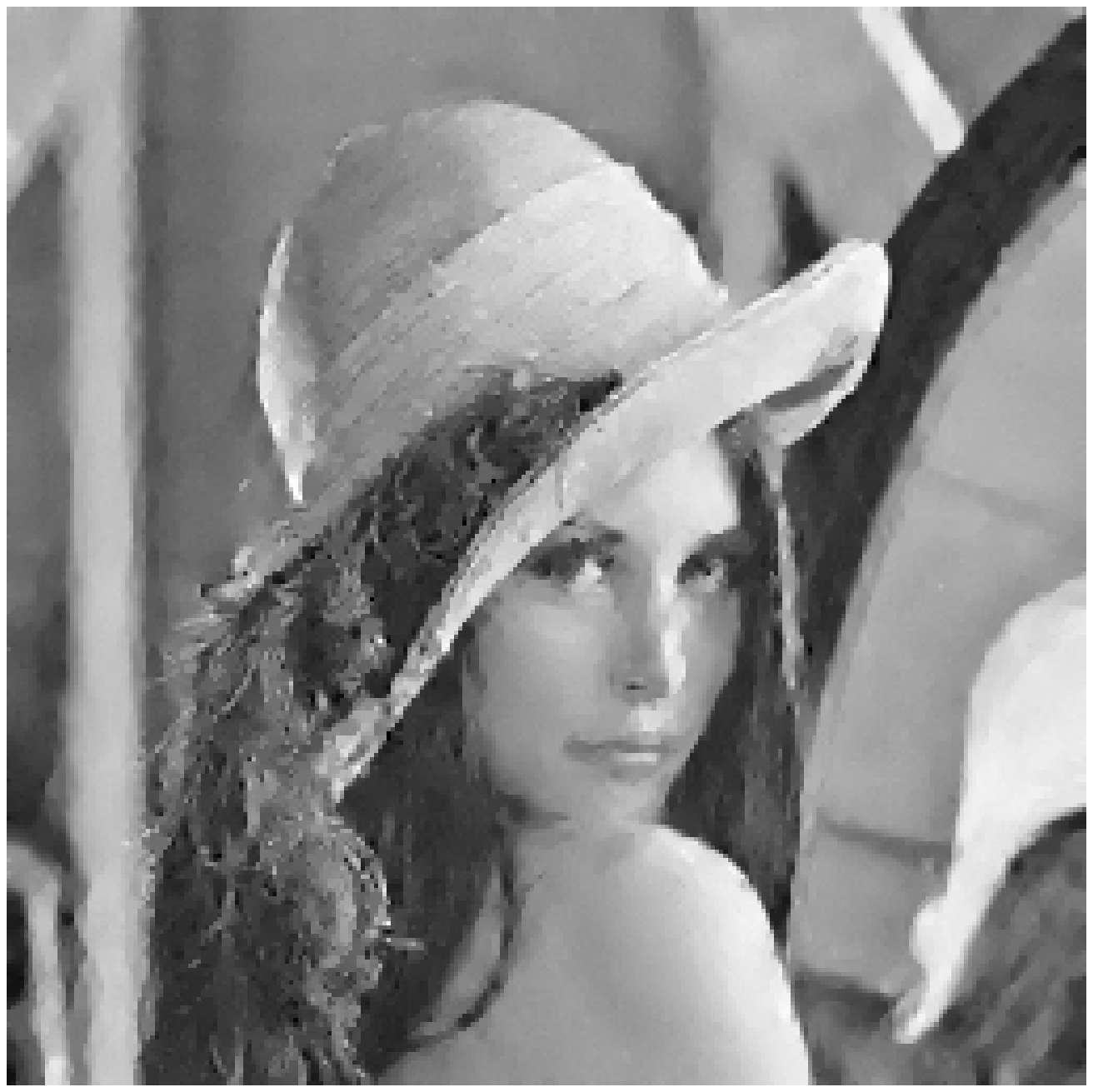}}
  \hspace{0pt}
  \subfigure[NLTV-BOS, PSNR=26.83dB, iter=15, CPU time=824.55s]{
    \label{fig:subfig:e} 
    \includegraphics[width=2.0in,clip]{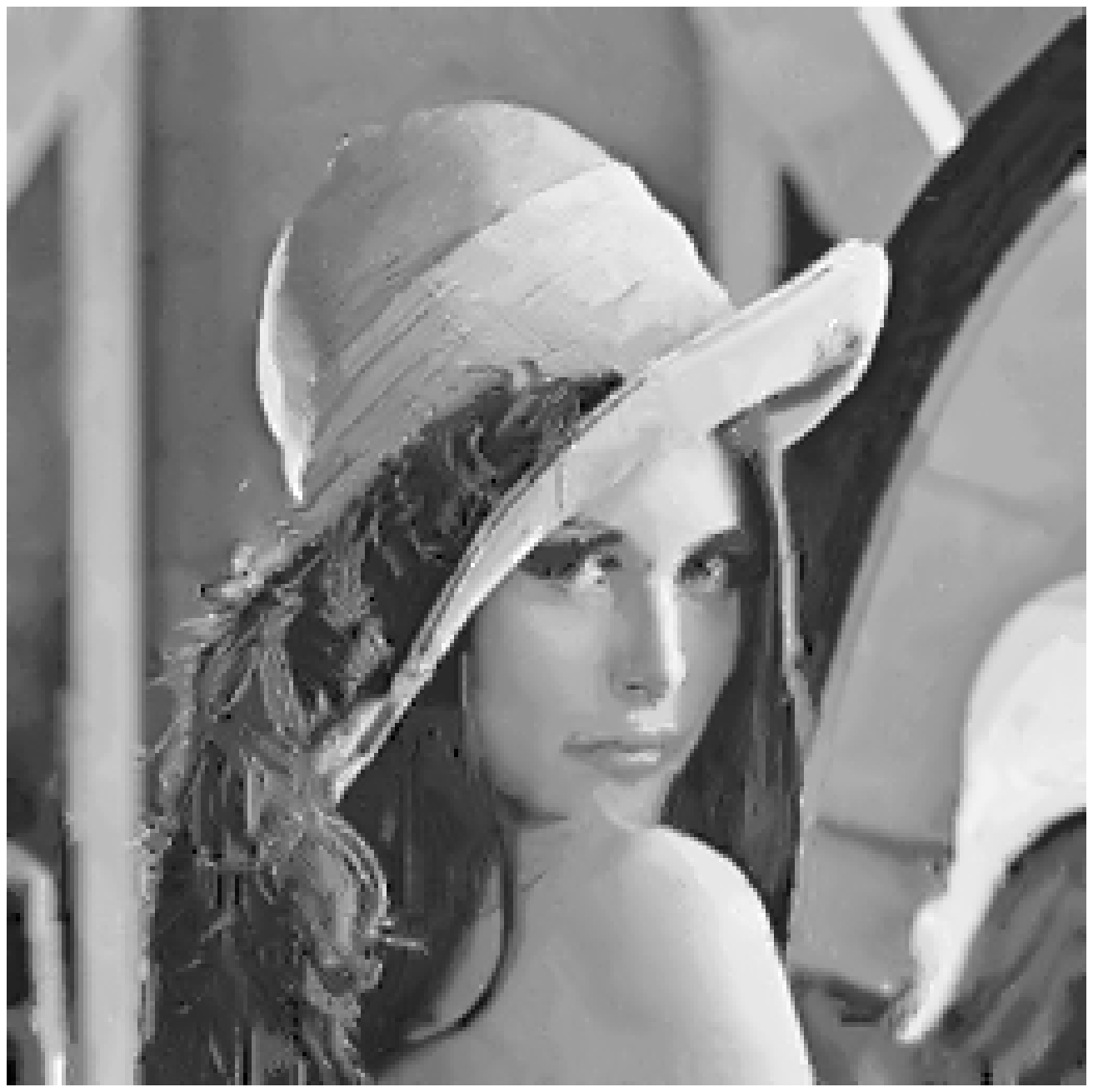}}
  \hspace{0pt}
  \subfigure[NLTV-Algorithm 1, PSNR=26.82dB, iter=25, CPU time=233.81s]{
    \label{fig:subfig:f} 
    \includegraphics[width=2.0in,clip]{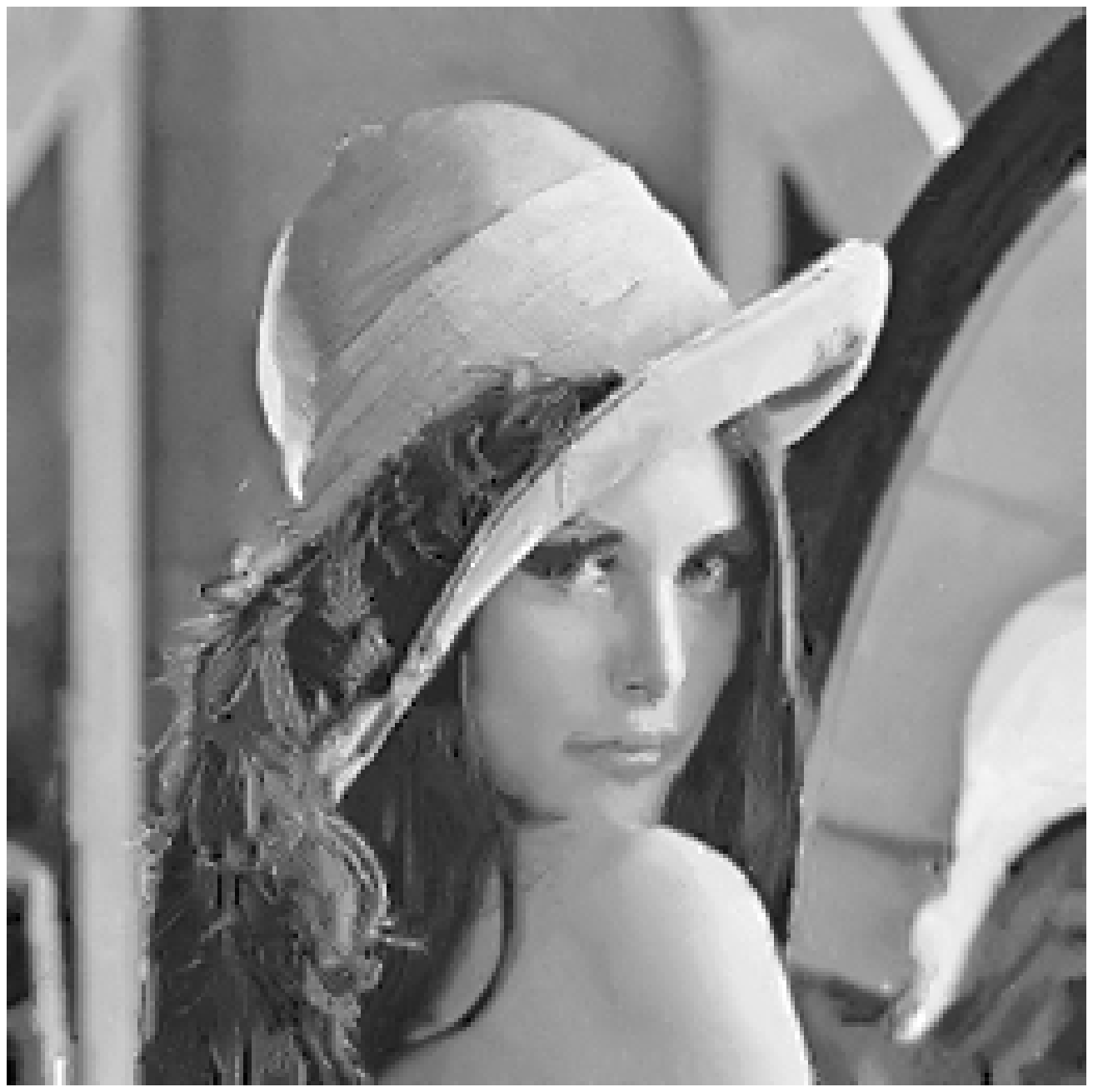}}\\
  \hspace{0pt}
  \label{fig:subfig} 
\caption{Random loss: $50\%$ random chosen frequencies. (a)Received
image, (b)restored image by applying the nearest neighbor
interpolation on the LL subband, (c)the result by BOS algorithm with
TV, (d)the result by Algorithm 1 with TV, (e)the result by BOS
algorithm with NL-TV, (f)the result by Algorithm 1 with NL-TV.}
\end{figure}

\begin{figure}
  \centering
  \subfigure[]{
    \label{fig:subfig:a} 
    \includegraphics[width=2.0in,clip]{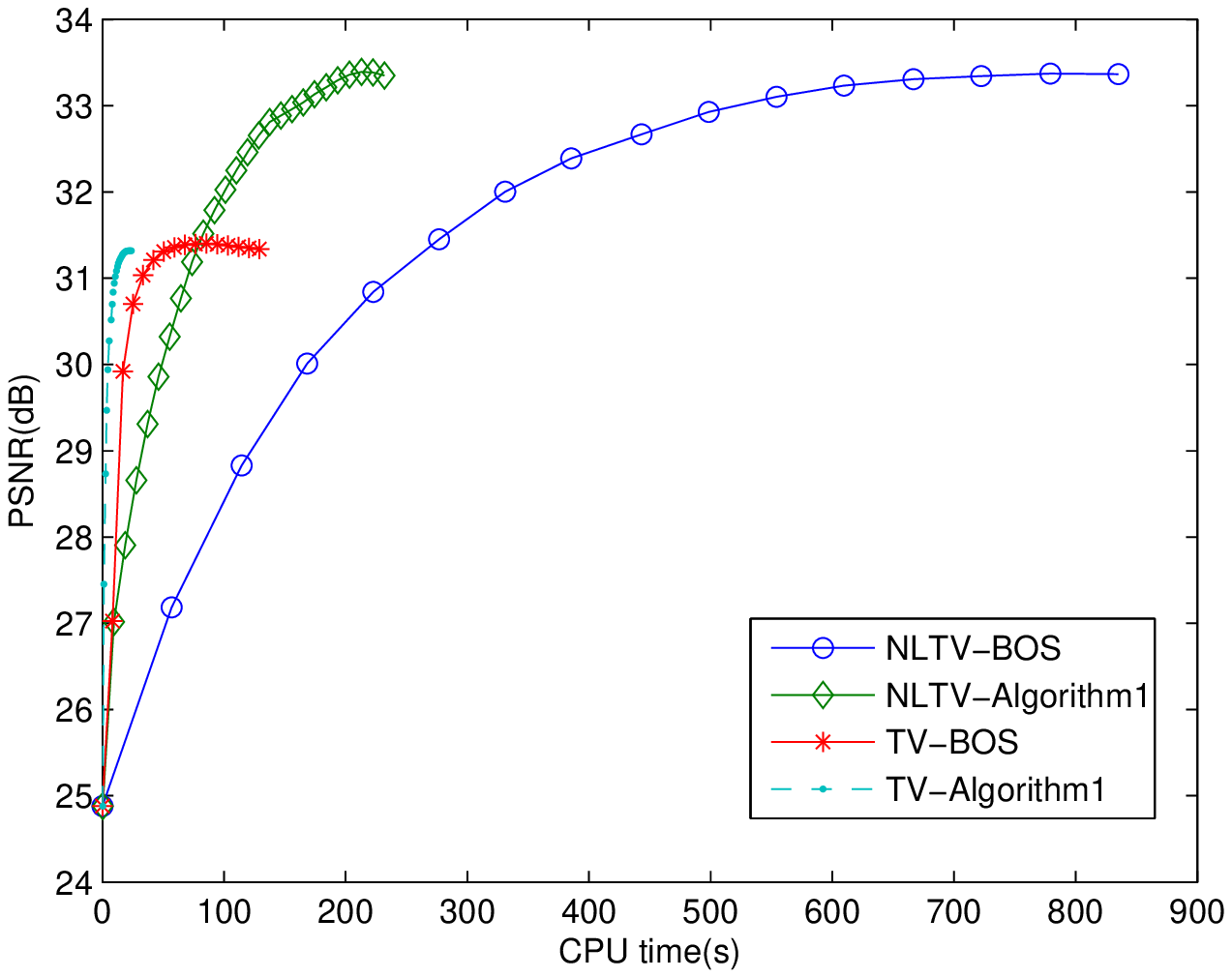}}
  \hspace{0pt}
  \subfigure[]{
    \label{fig:subfig:b} 
    \includegraphics[width=2.03in,clip]{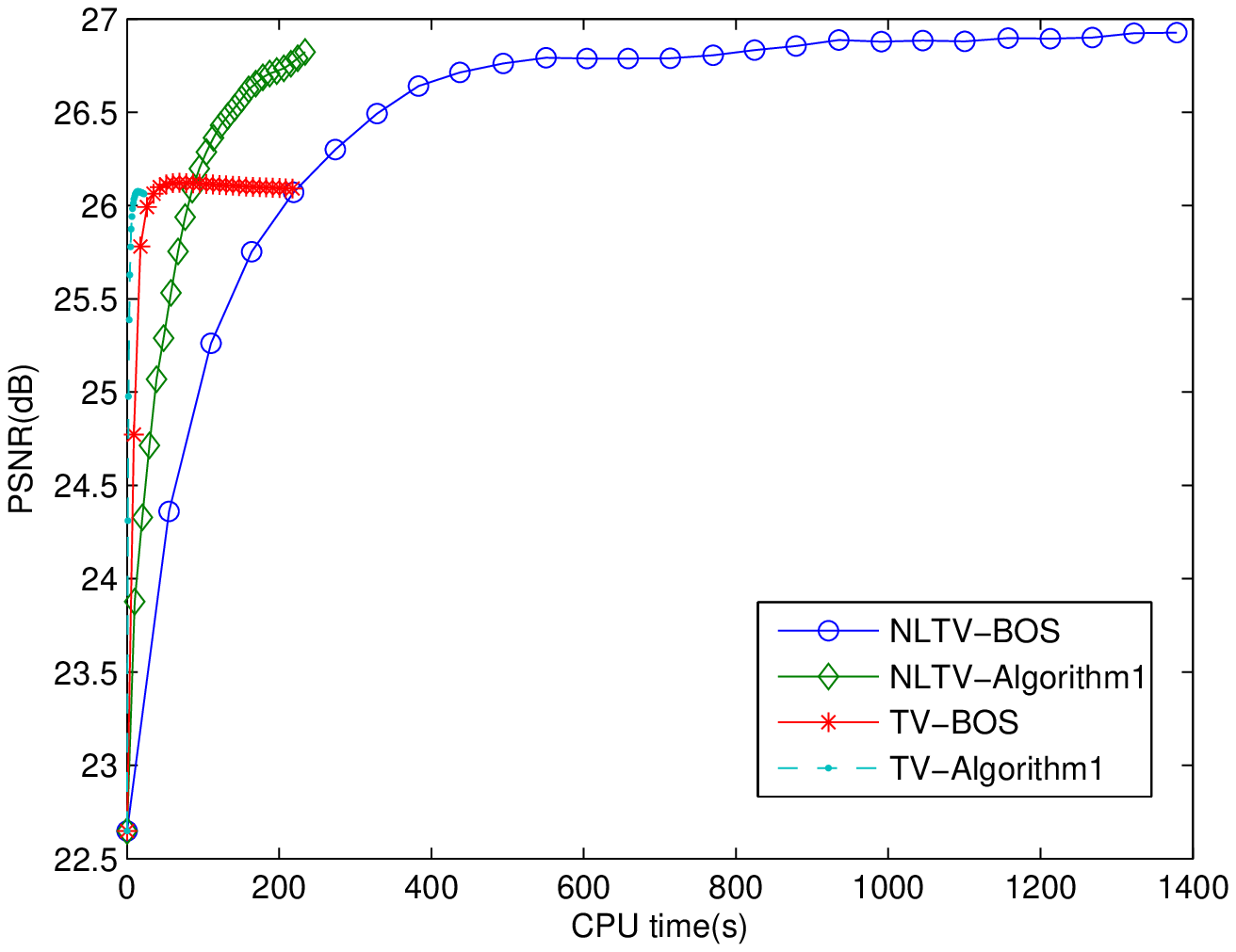}}
  \hspace{0pt}
  \label{fig:subfig} 
\caption{Evolution of PSNR against CPU time with the Lena image.
(a)The evolution curves of different methods for the received image
in Figure 4, (b)the evolution curves of different methods for the
received image in Figure 5.}
\end{figure}

Next, the images Cameraman and GoldHill are used to test both
algorithms with the TV regularization term. The four cases of
wavelet coefficients missing: the whole LH and HL loss, $50\%$
random loss keeping all low-low subband frequencies, $30\%$ random
loss are considered and the corresponding PSNR values and CPU time
are shown in Table 1, where '(H)' denotes only high frequencies are
randomly lost. The maximum outer iterations are set as 15 for the
first three cases, and set as 25 for the last case. It is observed
that our method is more efficient than the BOS algorithm in the CPU
time though the PSNR values are more or less the same. Figure 7-8
shows the restored results of both images Cameraman and GoldHill
with $30\%$ of their wavelet coefficients are lost randomly. We
observe that the proposed algorithm requires fewer computational
time than the BOS method to achieve the similar recovery effect.

\begin{table} [htbp]
\centering \caption{PSNR and CPU time for different methods}
\scalebox{1.0}{
\begin{tabular}{|c|c|c|c|c|c|}
  \hline
  Image & missing case & \multicolumn{2}{c|}{BOS algorithm} & \multicolumn{2}{c|}{Our method} \\
  \cline{3-6}
     &  & PSNR(dB) & Time(s) & PSNR(dB) & Time(s)  \\
  \hline
     & HL loss & 35.57 & 127.67 & 35.54 & 13.89  \\
  \cline{2-6}
    Cameraman & LH loss & 35.98 & 127.61 & 36.19 & 14.30 \\
  \cline{2-6}
     & 50$\%$ loss (H) & 24.88 & 126.23 & 24.84 & 14.12 \\
  \cline{2-6}
     & 30$\%$ loss & 27.76 & 212.27 & 27.55 & 22.33  \\
  \hline
     & HL loss & 32.13 & 124.06 & 32.44 & 14.79  \\
  \cline{2-6}
    GoldHill & LH loss & 31.87 & 125.55 & 32.22 & 14.18 \\
  \cline{2-6}
     & 50$\%$ loss (H) & 24.95 & 127.14 & 24.98 & 13.39 \\
  \cline{2-6}
     & 30$\%$ loss & 26.25 & 208.59 & 26.33 & 22.0  \\
  \hline
\end{tabular}}
\end{table}

\begin{figure}
  \centering
  \subfigure[Received image]{
    \label{fig:subfig:a} 
    \includegraphics[width=2.0in,clip]{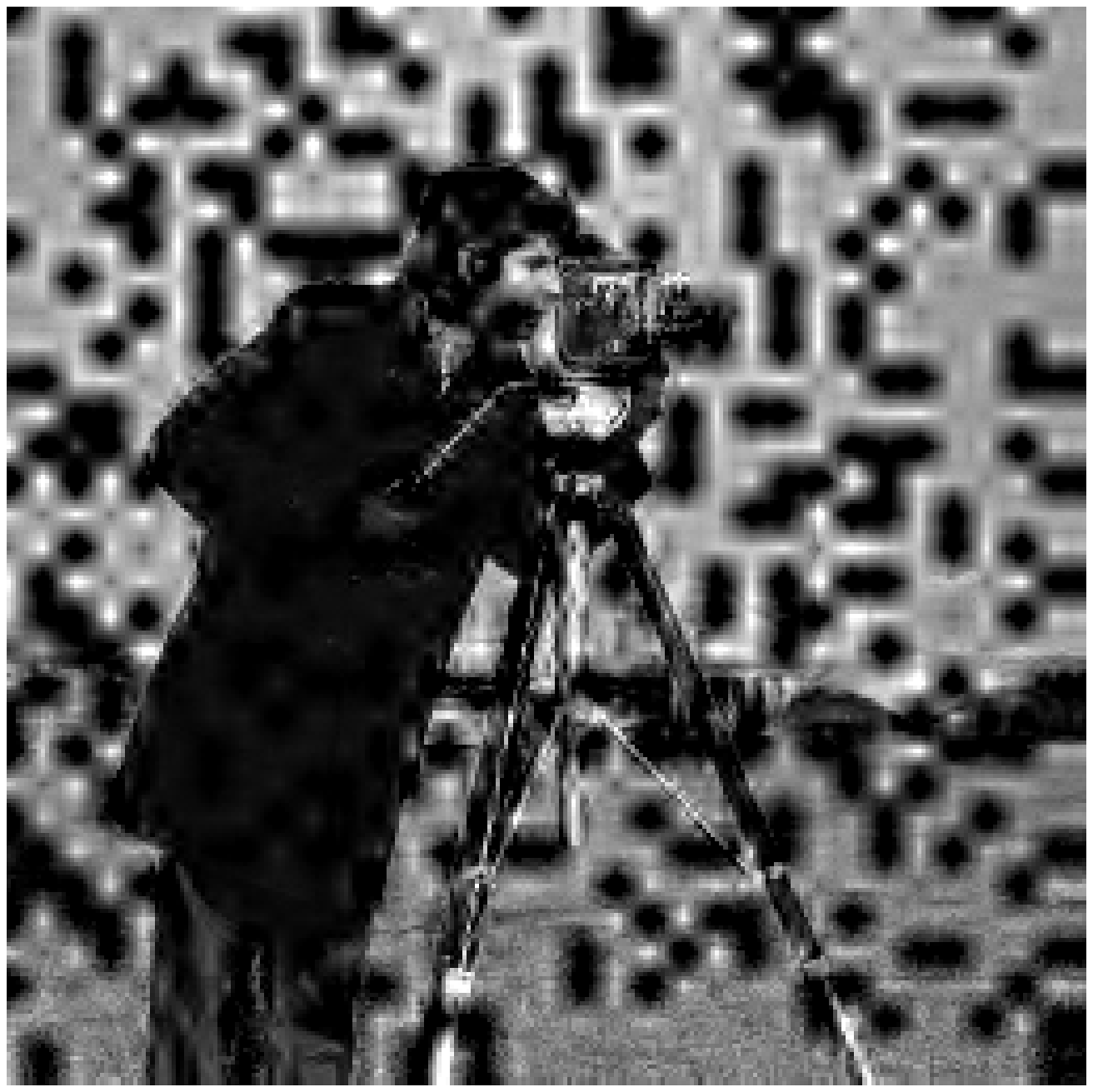}}
  \hspace{0pt}
  \subfigure[Interpolated image]{
    \label{fig:subfig:b} 
    \includegraphics[width=2.0in,clip]{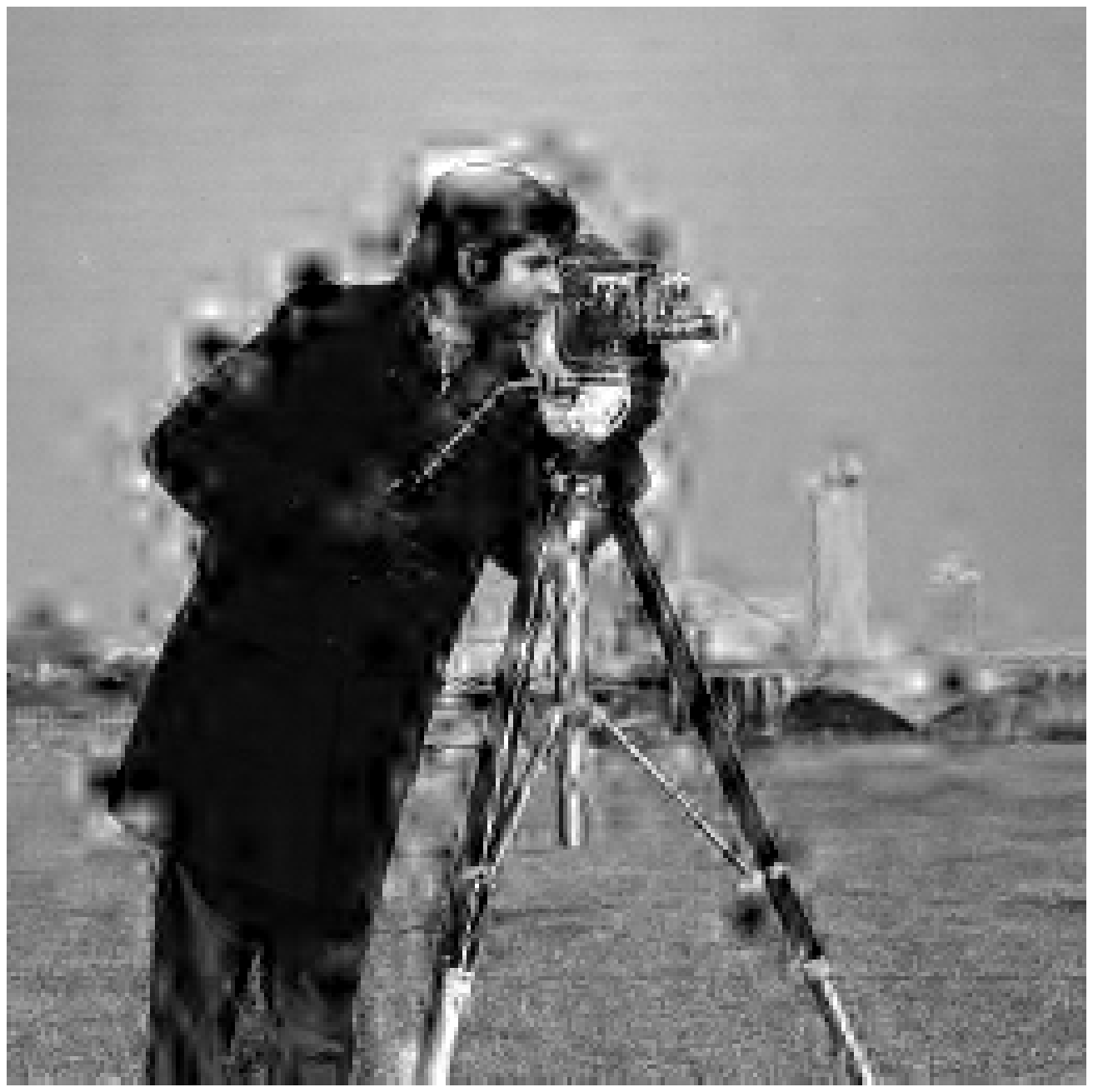}}\\
  \hspace{0pt}
  \subfigure[TV-BOS]{
    \label{fig:subfig:c} 
    \includegraphics[width=2.0in,clip]{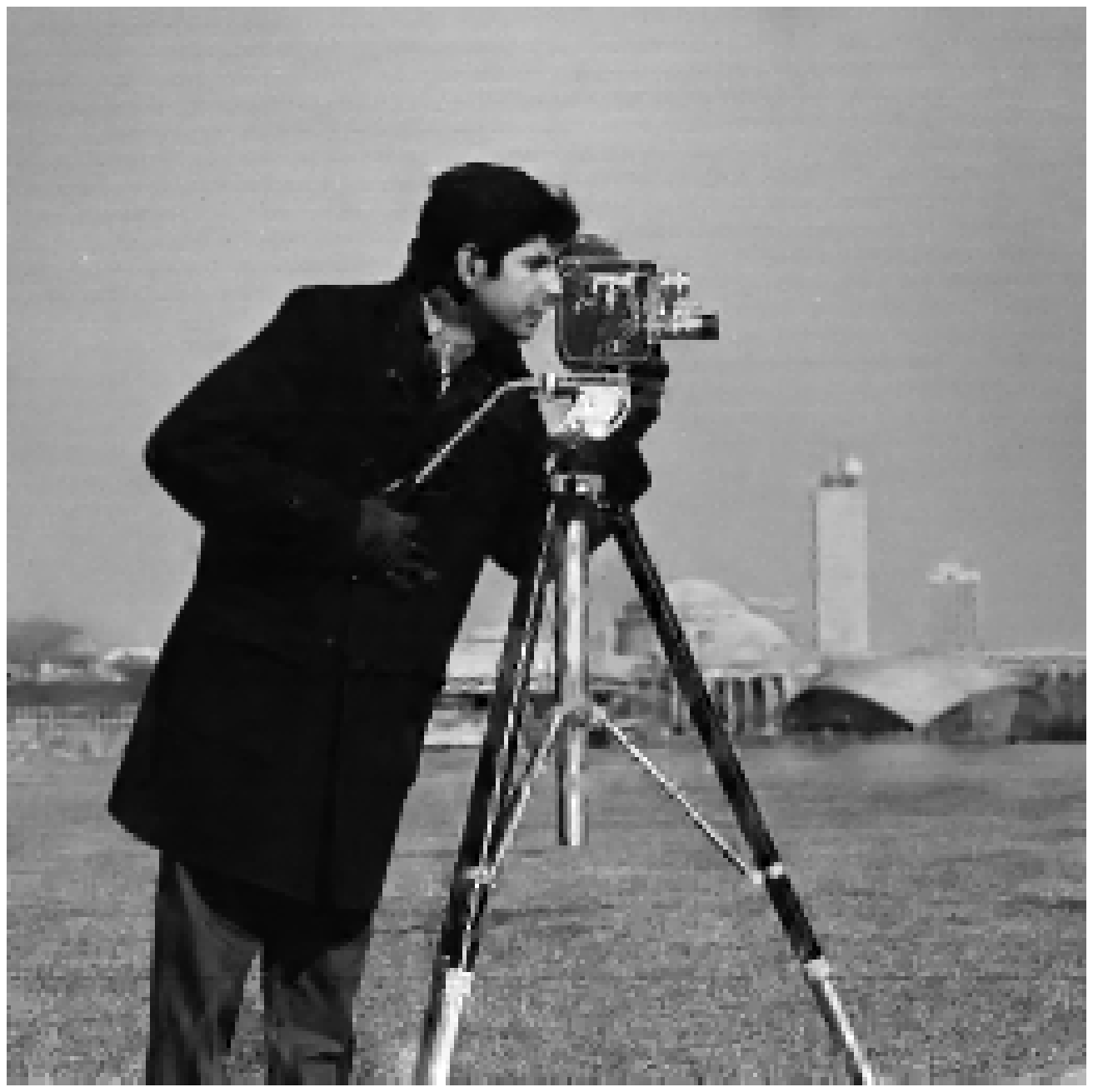}}
  \hspace{0pt}
  \subfigure[TV-Algorithm 1]{
    \label{fig:subfig:d} 
    \includegraphics[width=2.0in,clip]{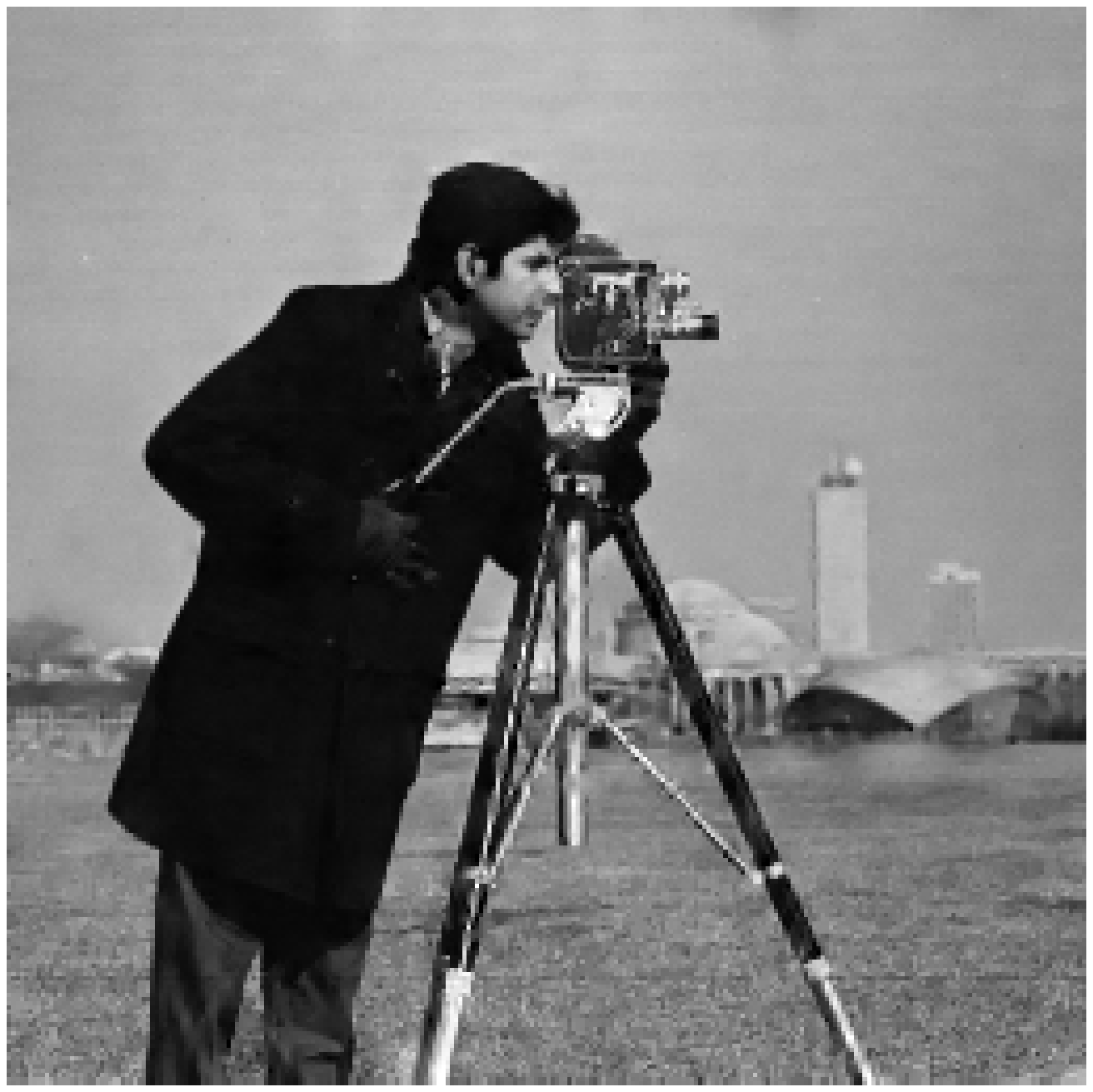}}
  \hspace{0pt}
  \label{fig:subfig} 
\caption{Random loss: $70\%$ random chosen frequencies. (a)Received
image, (b)restored image by applying the nearest neighbor
interpolation on the LL subband, (c)the result by BOS algorithm with
TV, (d)the result by Algorithm 1 with TV.}
\end{figure}

\begin{figure}
  \centering
  \subfigure[Received image]{
    \label{fig:subfig:a} 
    \includegraphics[width=2.0in,clip]{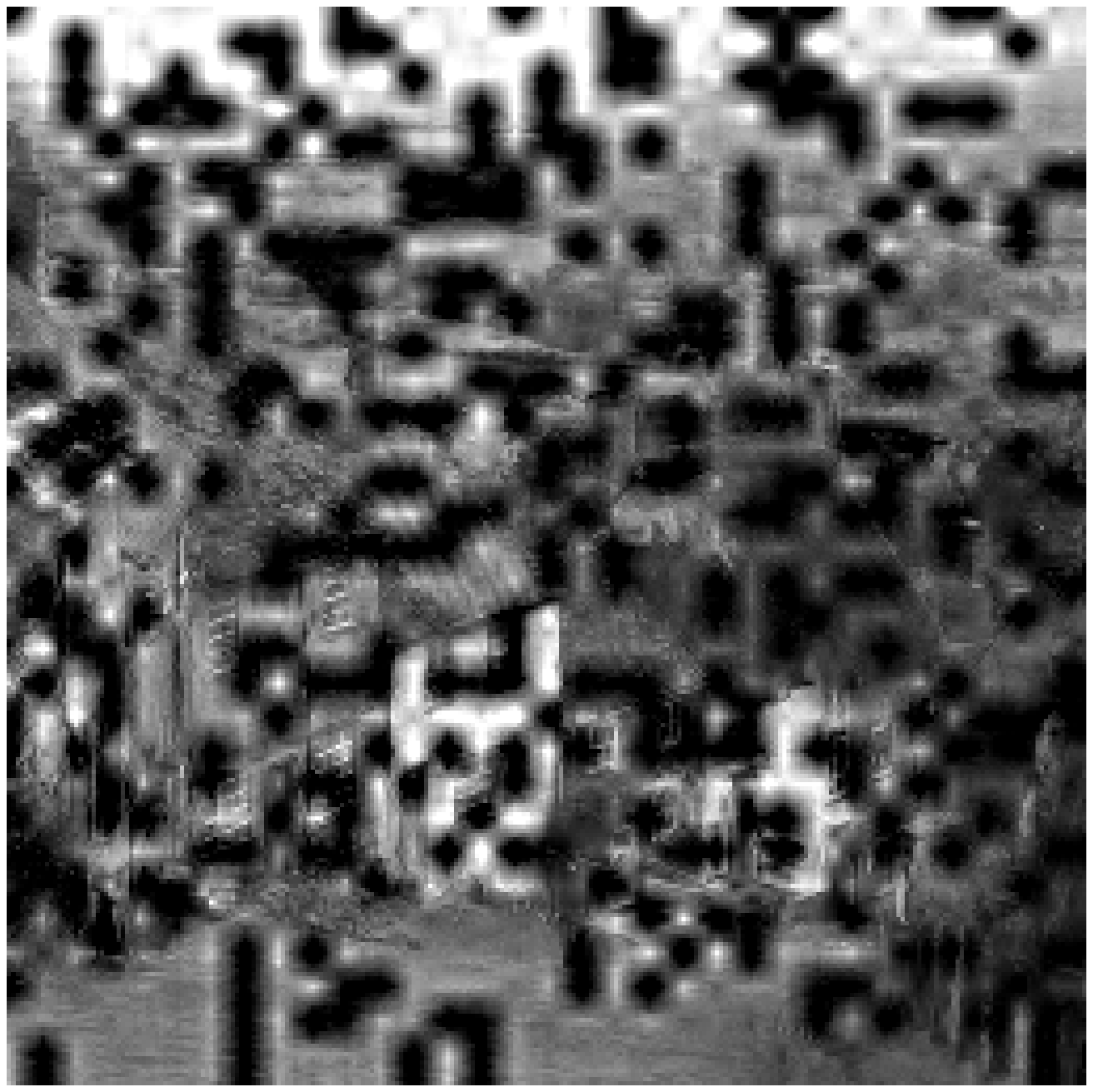}}
  \hspace{0pt}
  \subfigure[Interpolated image]{
    \label{fig:subfig:b} 
    \includegraphics[width=2.0in,clip]{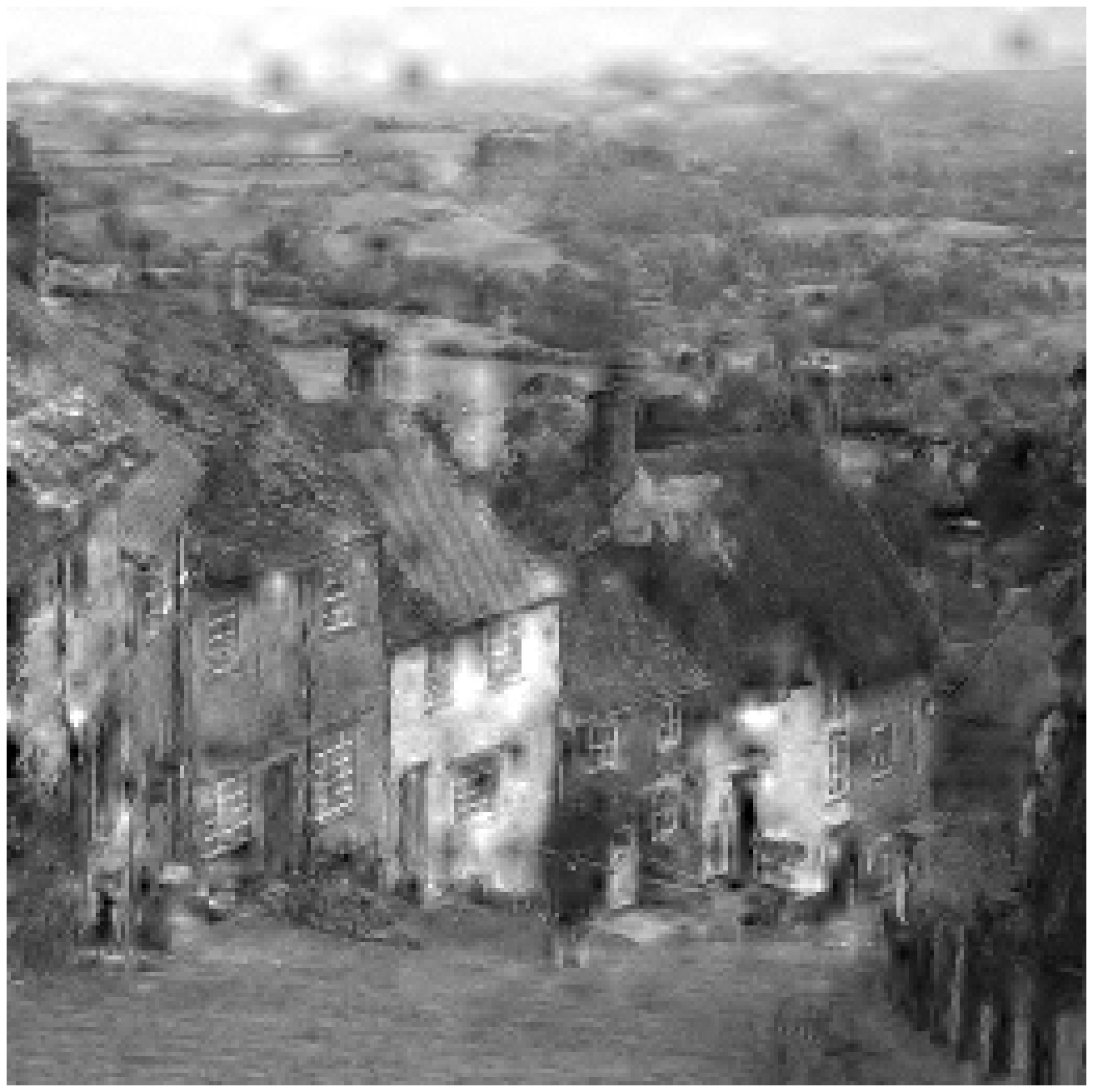}}\\
  \hspace{0pt}
  \subfigure[TV-BOS]{
    \label{fig:subfig:c} 
    \includegraphics[width=2.0in,clip]{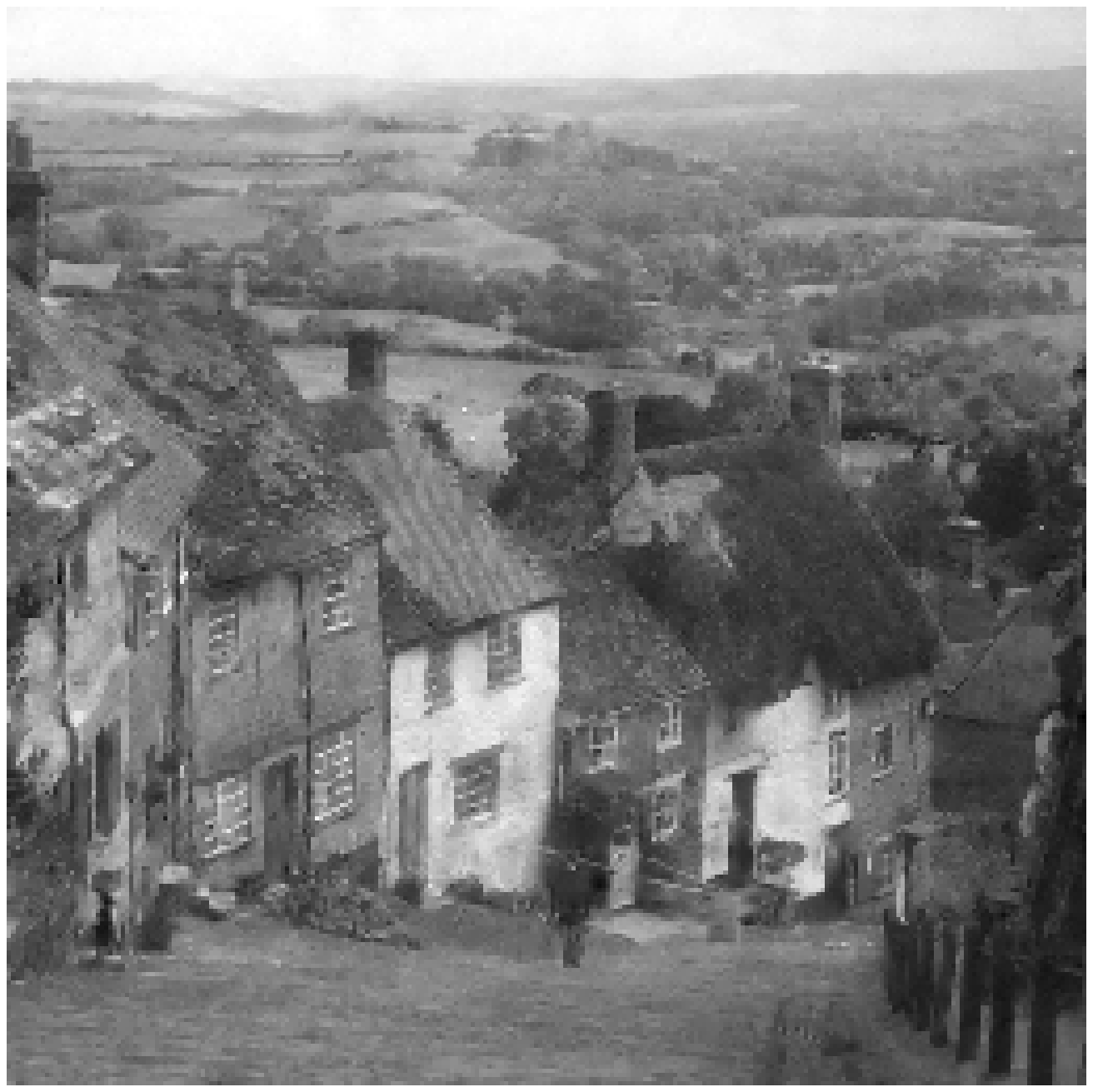}}
  \hspace{0pt}
  \subfigure[TV-Algorithm 1]{
    \label{fig:subfig:d} 
    \includegraphics[width=2.0in,clip]{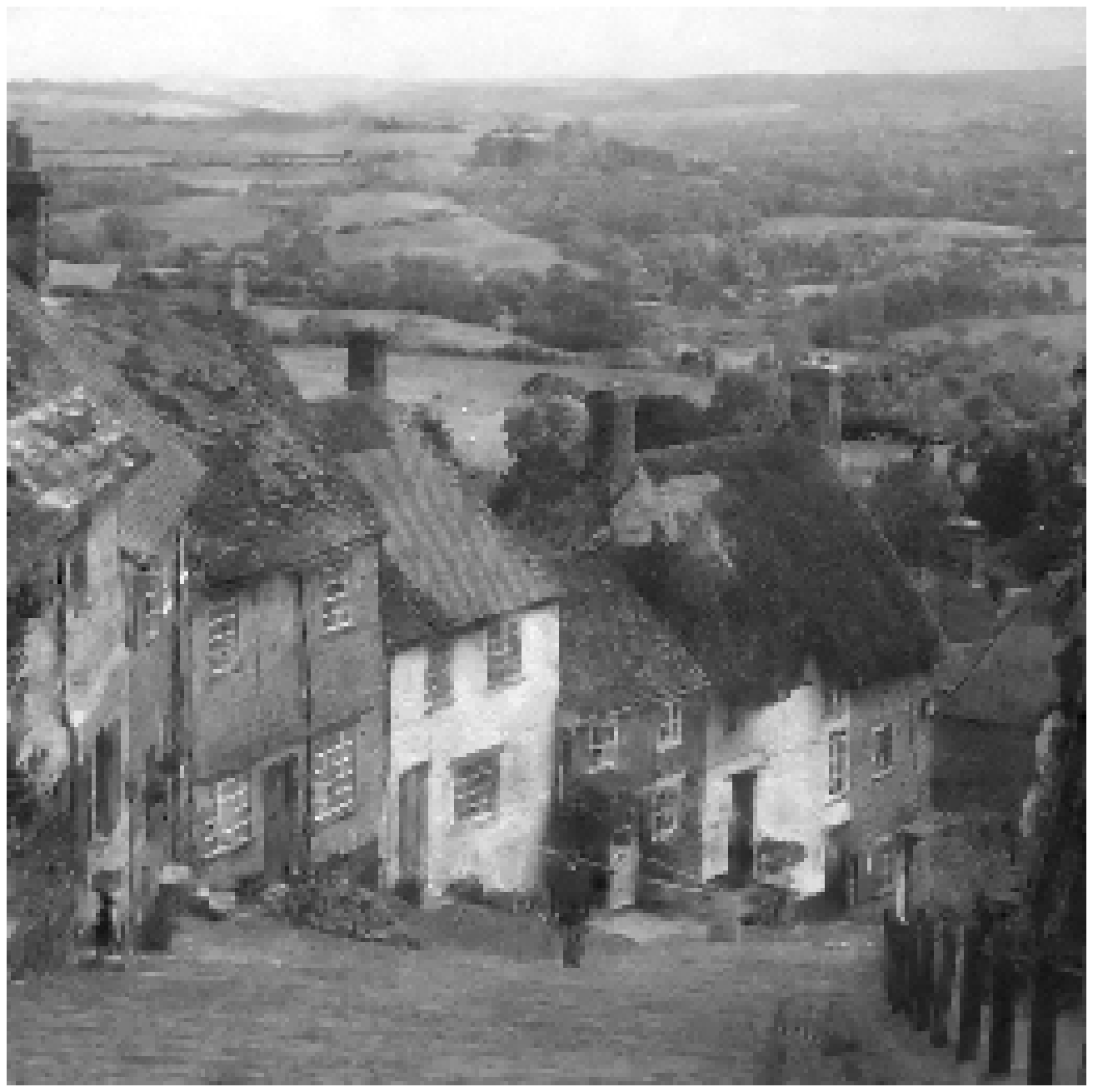}}
  \hspace{0pt}
  \label{fig:subfig} 
\caption{Random loss: $70\%$ random chosen frequencies. (a)Received
image, (b)restored image by applying the nearest neighbor
interpolation on the LL subband, (c)the result by BOS algorithm with
TV, (d)the result by Algorithm 1 with TV.}
\end{figure}

Finally, we test the case when there is white Gaussian noise present
in the random loss data. Similarly to the noise-free case, we use
the interpolated image as the initial guess for the BOS algorithm
and the known component $f_{0}$ in our method. Two images barba128
(size of $128\times 128$) and Cameraman are used for this test. Due
to existence of the noise we use $f^{k}$ in Algorithm 1 as the final
result, and the standard deviation $\sigma$ of the noise is used to
replace the upper bound $10^{-5}$ in the stopping criterion. Figure
9 shows the results of the noisy image barba128 generated by both
algorithms. It is obvious that our algorithm obtains the better
results with much less implementation time. The results of Cameraman
image corrupted by Gaussian noise with $\sigma=0.02$ are also shown
in Figure 10 (due to the NL-TV regularization cannot produce better
results for the image in this example, we don't consider it here). It
verifies the efficiency of the proposed algorithm once again.

\begin{figure}
  \centering
  \subfigure[Received image]{
    \label{fig:subfig:a} 
    \includegraphics[width=2.0in,clip]{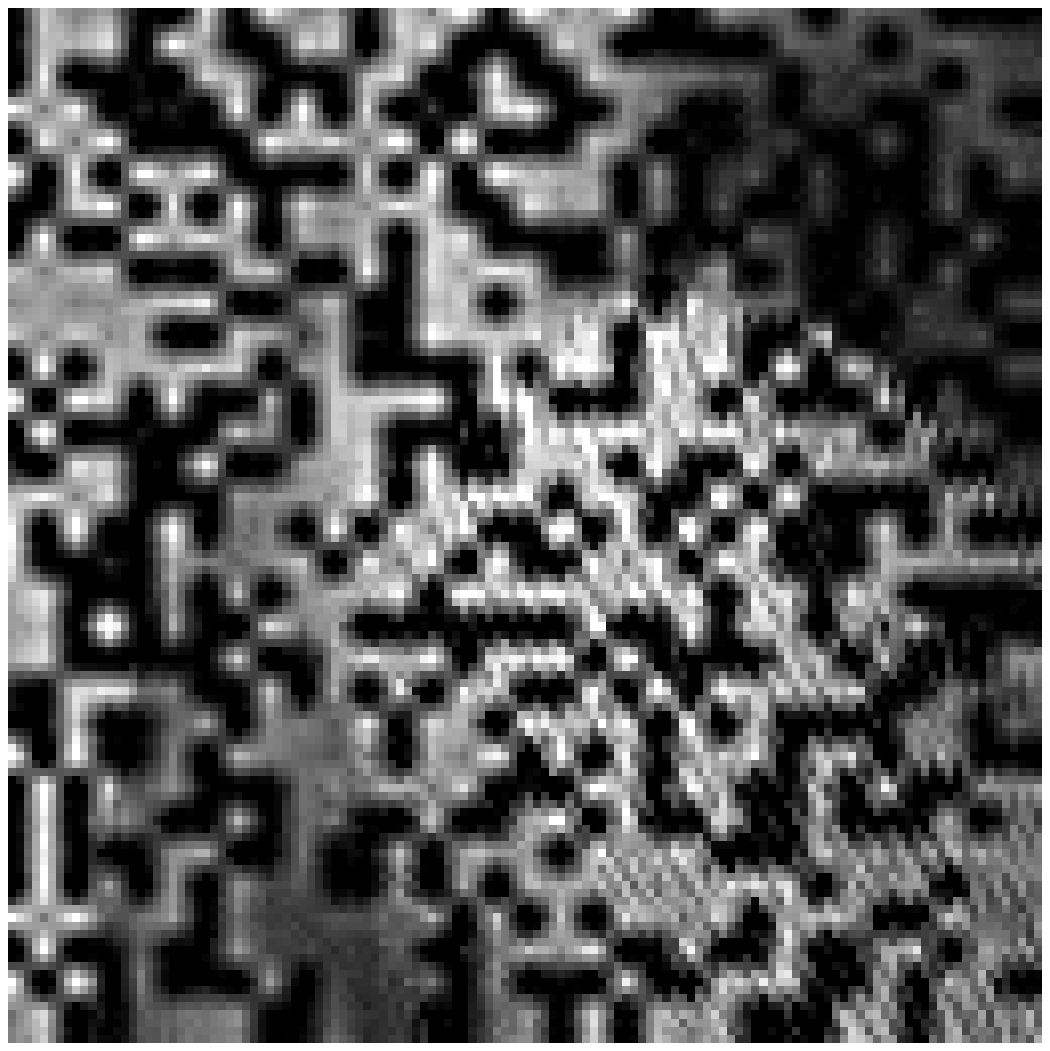}}
  \hspace{0pt}
  \subfigure[Interpolated image, PSNR=23.49dB]{
    \label{fig:subfig:b} 
    \includegraphics[width=2.0in,clip]{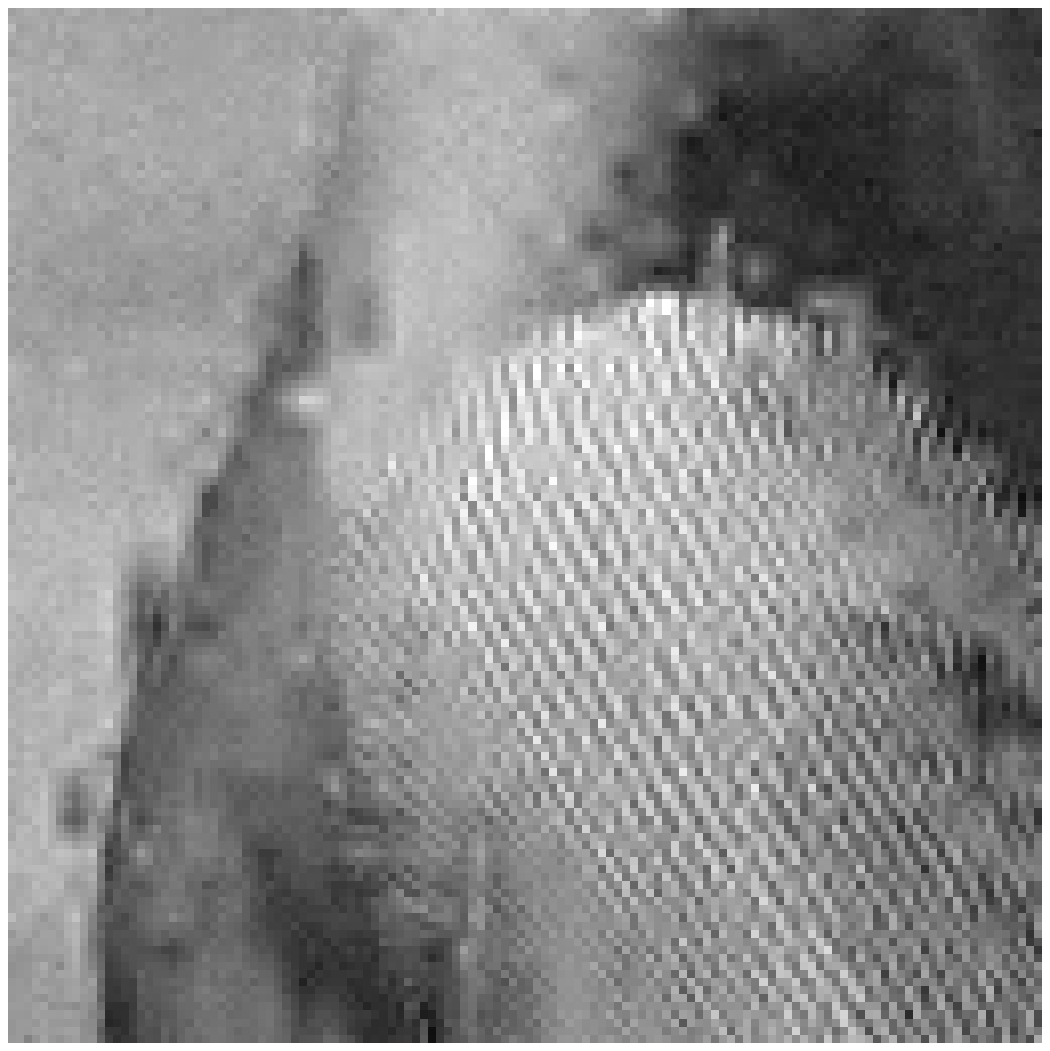}}
  \hspace{0pt}
  \subfigure[TV-BOS, PSNR=23.99dB, iter=5, CPU time=27.20s]{
    \label{fig:subfig:c} 
    \includegraphics[width=2.0in,clip]{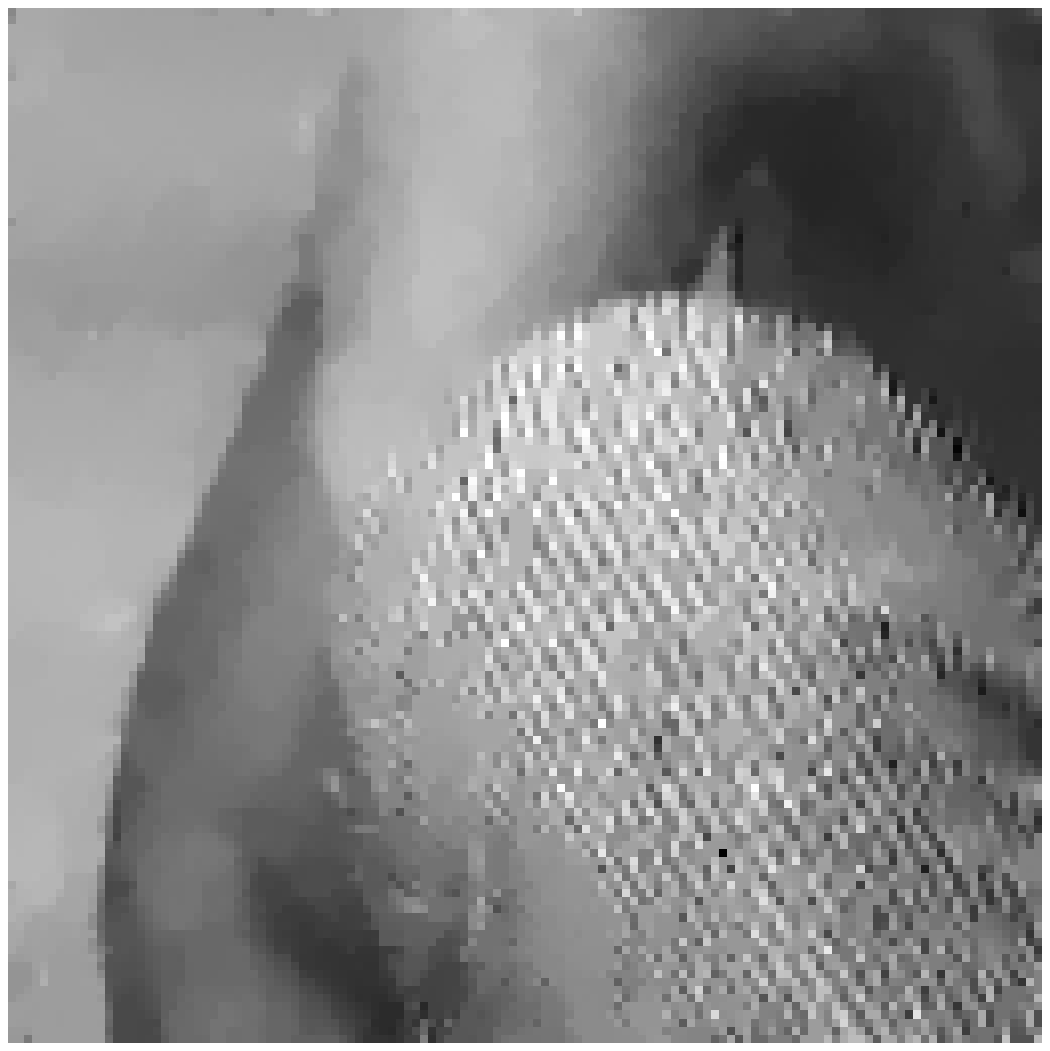}}
  \hspace{0pt}
  \subfigure[TV-Algorithm 1, PSNR=24.14dB, iter=3, CPU time=0.64s]{
    \label{fig:subfig:d} 
    \includegraphics[width=2.0in,clip]{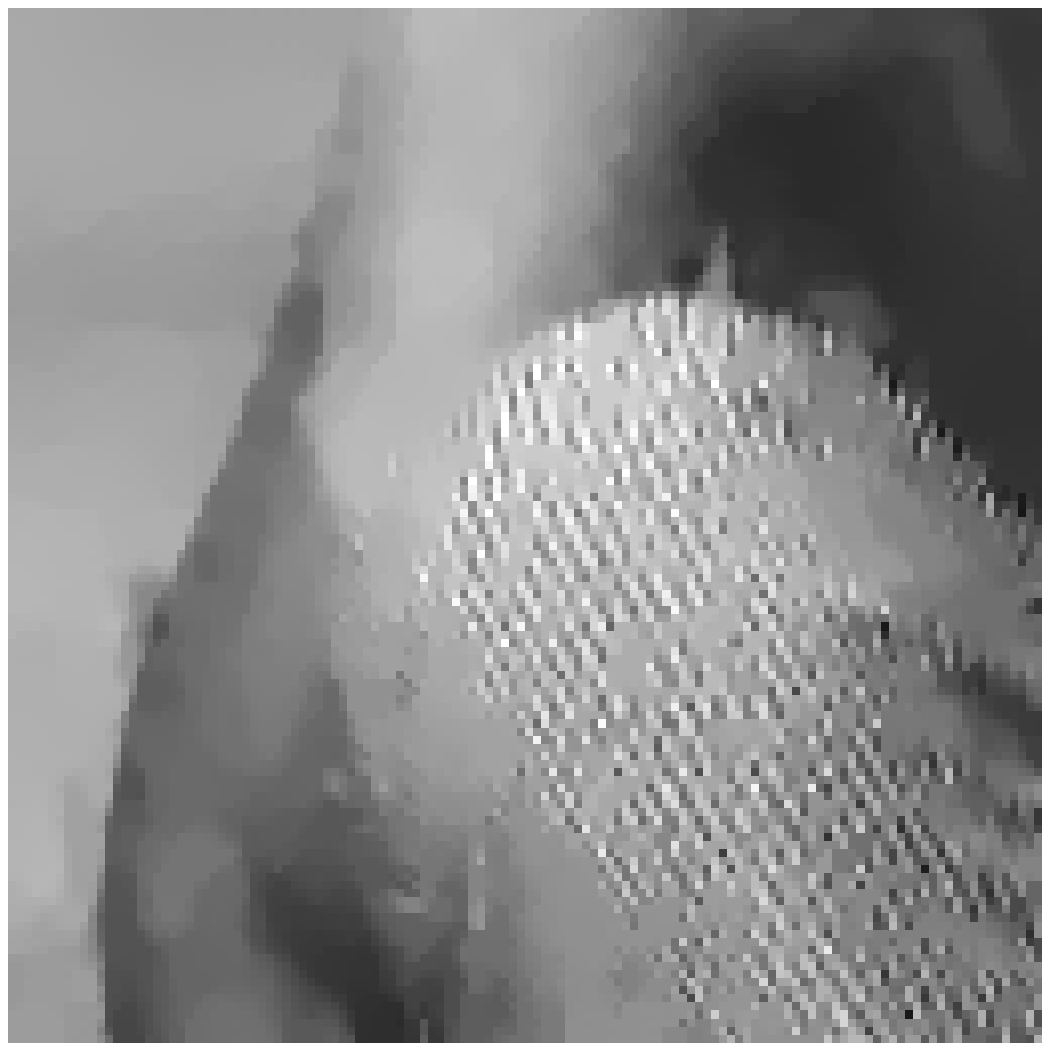}}
  \hspace{0pt}
  \subfigure[NLTV-BOS, PSNR=27.94dB, iter=15, CPU time=486.95s]{
    \label{fig:subfig:e} 
    \includegraphics[width=2.0in,clip]{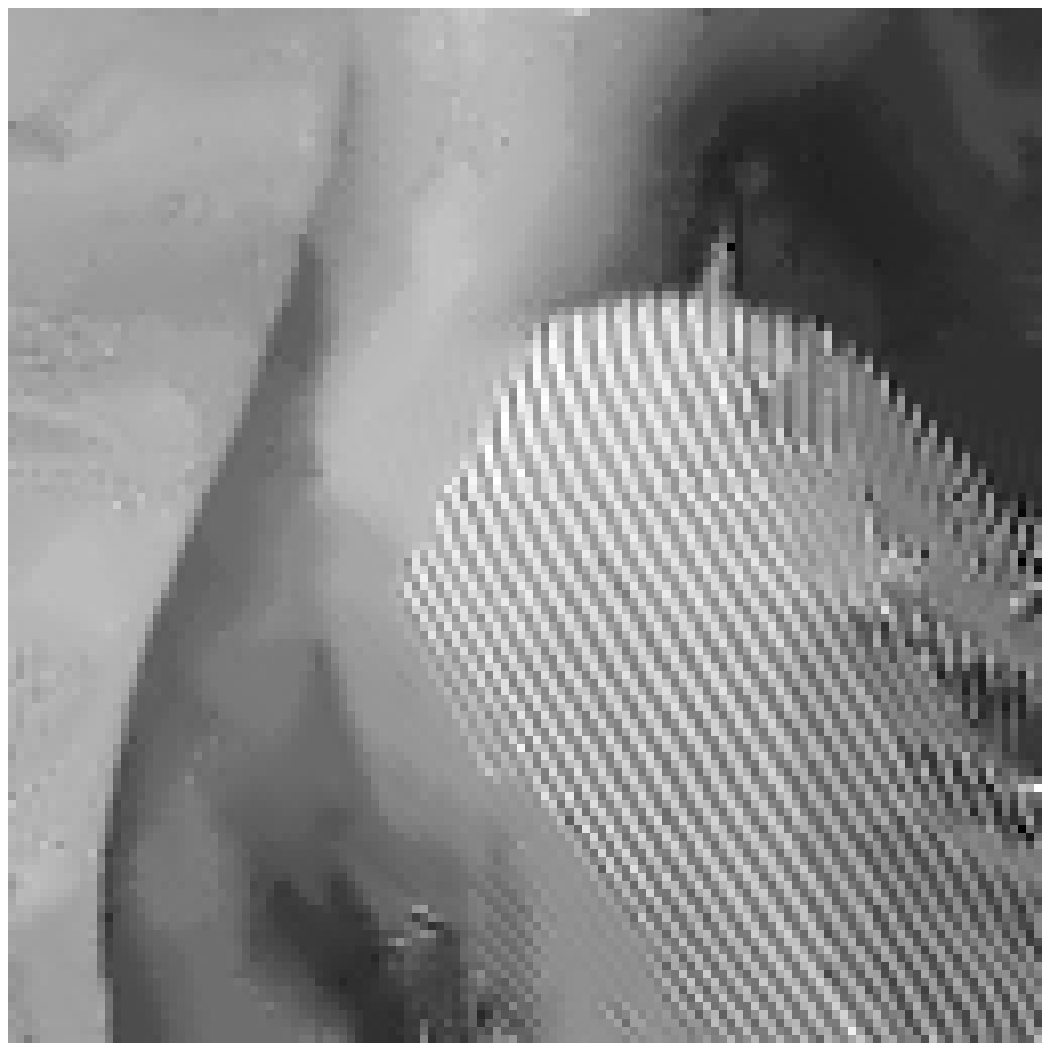}}
  \hspace{0pt}
  \subfigure[NLTV-Algorithm 1, PSNR=28.35dB, iter=8, CPU time=18.16s]{
    \label{fig:subfig:f} 
    \includegraphics[width=2.0in,clip]{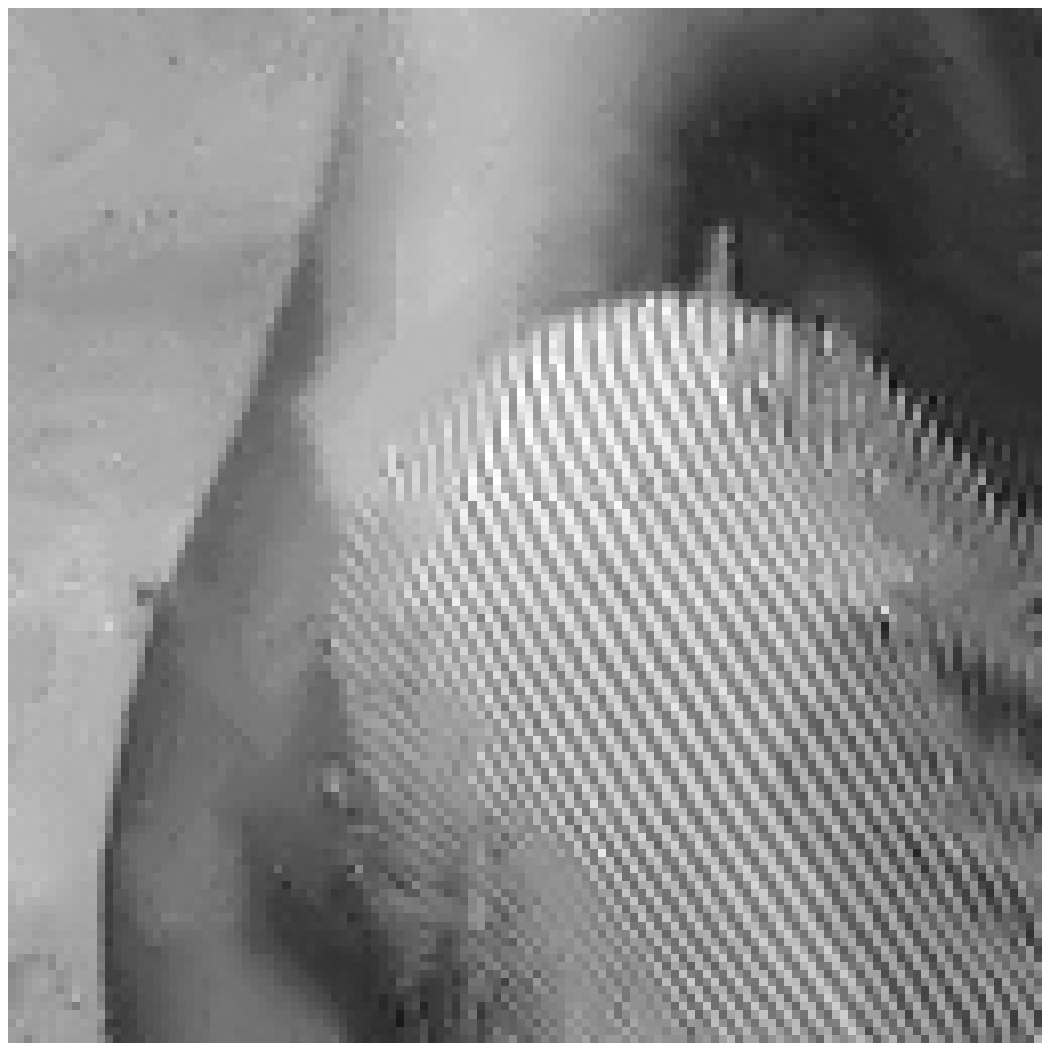}}
  \hspace{0pt}
  \label{fig:subfig} 
\caption{Random loss with noise: $60\%$ random chosen frequencies
with noise level $\sigma=0.02$. (a)Received image, (b)restored image
by applying the nearest neighbor interpolation on the LL subband,
(c)the result by BOS algorithm with TV, (d)the result by Algorithm 1
with TV, (e)the result by BOS algorithm with NL-TV, (f)the result by
Algorithm 1 with NL-TV}
\end{figure}

\begin{figure}
  \centering
  \subfigure[Received image]{
    \label{fig:subfig:a} 
    \includegraphics[width=2.0in,clip]{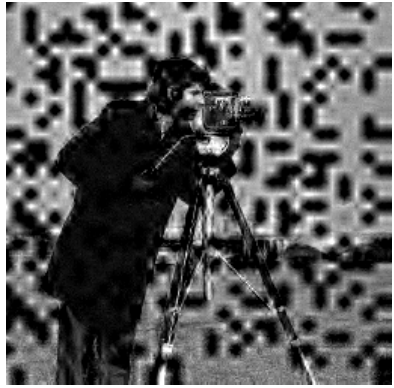}}
  \hspace{0pt}
  \subfigure[Interpolated image, PSNR=20.49dB]{
    \label{fig:subfig:b} 
    \includegraphics[width=2.0in,clip]{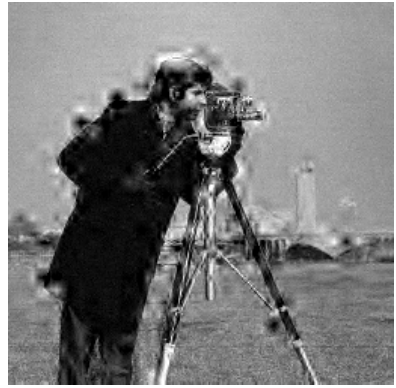}}\\
  \hspace{0pt}
  \subfigure[TV-BOS, PSNR=25.89dB, iter=3, CPU time=28.86s]{
    \label{fig:subfig:c} 
    \includegraphics[width=2.0in,clip]{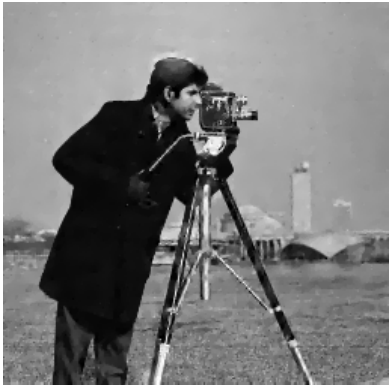}}
  \hspace{0pt}
  \subfigure[TV-Algorithm 1, PSNR=25.89dB, iter=10, CPU time=9.23s]{
    \label{fig:subfig:d} 
    \includegraphics[width=2.0in,clip]{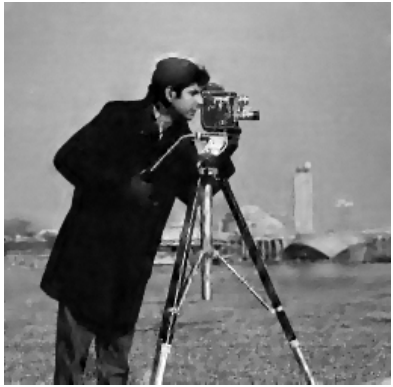}}
  \hspace{0pt}
  \label{fig:subfig} 
\caption{Random loss with noise: $70\%$ random chosen frequencies
with noise level $\sigma=0.02$. (a)Received image, (b)restored image
by applying the nearest neighbor interpolation on the LL subband,
(c)the result by BOS algorithm with TV, (d)the result by Algorithm 1
with TV.}
\end{figure}

\section{Conclusion}\label{sec5}

In this article, inspired by the idea of image decomposition we
propose a novel wavelet domain inpainting model, and present a fast
iterative algorithm based on the split-Bregman method to solve the
corresponding minimization problem. The convergence properties of
the iterative algorithms have also been researched. Numerical
examples demonstrate that the proposed method is more efficient than
the previous works, especially in the computational time.

\end{document}